\documentclass[letterpaper]{article} 
\usepackage[submission]{aaai23}  
\usepackage{times}  
\usepackage{helvet}  
\usepackage{courier}  
\usepackage[hyphens]{url}  
\usepackage{graphicx} 
\usepackage{natbib}  
\usepackage{caption} 
\usepackage{algorithm}
\usepackage{algorithmic}

\usepackage{newfloat}
\usepackage{listings}
\DeclareCaptionStyle{ruled}{labelfont=normalfont,labelsep=colon,strut=off} 
\floatstyle{ruled}
\newfloat{listing}{tb}{lst}{}
\floatname{listing}{Listing}

\usepackage{amssymb}
\usepackage{booktabs}
\usepackage{amsmath}
\usepackage{multirow}
\usepackage{makecell}
\usepackage{balance}
\usepackage{threeparttable}
\newtheorem{theorem}{Theorem}[section]

\newtheorem{definition}[theorem]{Definition}
\newtheorem{assumption}[theorem]{Assumption}

\newcommand*{\Scale}[2][4]{\scalebox{#1}{$#2$}}

\title{Learning Instrumental Variable from Data Fusion \\ for Treatment Effect Estimation}

\author{
    Anpeng Wu\textsuperscript{\rm 1},
    Kun Kuang\textsuperscript{\rm 1}\thanks{Corresponding author.}, 
    Ruoxuan Xiong\textsuperscript{\rm 2},
    Minqing Zhu\textsuperscript{\rm 1},
    Yuxuan Liu\textsuperscript{\rm 1},
    Bo Li\textsuperscript{\rm 3},
    Furui Liu\textsuperscript{\rm 4},
    Zhihua Wang\textsuperscript{\rm 5,6},
    Fei Wu\textsuperscript{\rm 1,5,6}
}
\affiliations{
    \textsuperscript{\rm 1} Department of Computer Science and Technology, Zhejiang University\\
    \textsuperscript{\rm 2} Department of Quantitative Theory and Methods, Emory University\\
    \textsuperscript{\rm 3} School of Economics and Management, Tsinghua University\\
    \textsuperscript{\rm 4} Huawei Noah's Ark lab, \textsuperscript{\rm 5} Shanghai AI Laboratory\\
    \textsuperscript{\rm 6} Shanghai Institute for Advanced Study of Zhejiang University
}

\usepackage{bibentry}

\begin{document}

\maketitle

\begin{abstract}
The advent of the big data era brought new opportunities and challenges to draw treatment effect in data fusion, that is, a mixed dataset collected from multiple sources (each source with an independent treatment assignment mechanism). Due to possibly omitted source labels and unmeasured confounders, traditional methods cannot estimate individual treatment assignment probability and infer treatment effect effectively. Therefore, we propose to reconstruct the source label and model it as a Group Instrumental Variable (GIV) to implement IV-based Regression for treatment effect estimation. In this paper, we conceptualize this line of thought and develop a unified framework (Meta-EM) to (1) map the raw data into a representation space to construct Linear Mixed Models for the assigned treatment variable; (2) estimate the distribution differences and model the GIV for the different treatment assignment mechanisms; and (3) adopt an alternating training strategy to iteratively optimize the representations and the joint distribution to model GIV for IV regression. Empirical results demonstrate the advantages of our Meta-EM compared with state-of-the-art methods. 
The code is available at https://github.com/anpwu/meta-em.
\end{abstract}

\section{Introduction}

%
Estimating the causal effects of treatment/exposure on the outcome of interest from the observation dataset is crucial for explanatory analysis and decision-making \cite{Relate5:pearl2009causal,kuang2020causal,li2020survey,zhang2021causerec,tian2022confoundergan}. In the presence of unmeasured confounders, assuming a fixed additive noise model (ANM), state-of-the-art (SOTA) approaches use an instrumental variable (IV) to implement a two-stage regression to reduce endogenous confounding bias in treatment effect estimation \cite{IVMethod1:hartford2017deepiv,IVMethod4:lin2019onesiv, IVMethod3:muandet2019dualiv,wu2022instrumental}. 
These methods are reliable when the pre-defined IV is a valid IV that only affects the outcome through its strong association with treatment options, called exclusion assumption.
Under these assumptions,  \citet{Identify2:angrist1996identification,Identify3:newey2003instrumental} verify that causal effects can be identified by exogenous IVs. 
In instrumental variable literature, researchers usually implement Randomized Controlled Trials (RCTs) to obtain exogenous IVs, such as Oregon health insurance experiment \cite{finkelstein2012oregon} and effects of military service on lifetime earnings \cite{angrist1990lifetime}, which are too expensive to be universally available.

\begin{figure}[t]
\centering
\includegraphics[width=0.95\columnwidth]{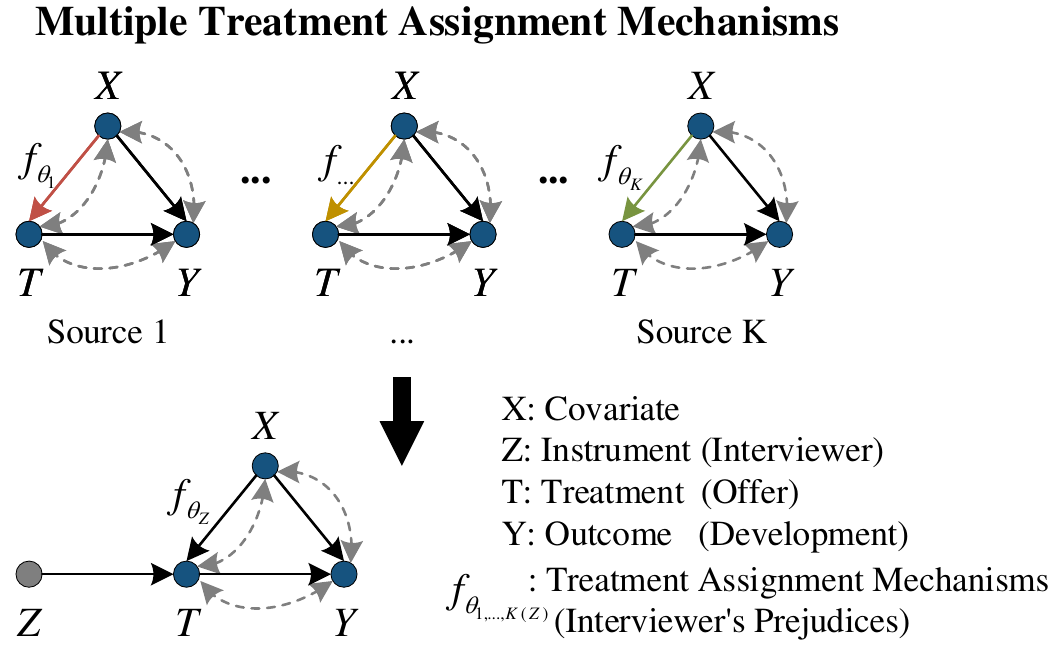} 
\caption{The causal diagram for mixed datasets from multiple sources, each source with an independent treatment assignment mechanism. Blue nodes denote observable variables, and gray nodes indicate latent variables. The arrows with different colors define different causal effects. The bidirected arrows encode unmeasured confounders.}
\label{fig1}
\end{figure}

%
With the advent of the big data era, a variety of observation databases collected from different sources have been established, which may contain the same treatment effect mechanism (from treatment to outcome) but different treatment assignment mechanisms (from covariates to treatment) \cite{bareinboim2016causal,hunermund2019causal}.
For instance, as shown in Fig. \ref{fig1}, in the study of treatment effect of individual offers (treatment $T$) on enterprise development (outcome $Y$) , different human resources (HR) interviewers (instrument $Z$) may assign different offer decisions to the same individual (covariate $X$) based on different evaluation strategies (assignment $f_{\theta_Z}$). In this case, candidates will be randomly assigned to different interviewers, each with different prejudices or opinions, to decide whether to give an offer or not \cite{pager2009bayesian}. Here, the omitted interviewer label (source label) can serve as a latent multi-valued IV, which only affects the outcome through its strong association with offer decisions \citep{IVSelection4:kuang2020ivy,CD2:rothenhausler2021anchor}. 
Such heterogeneous assignment mechanism is common and widespread in real applications, such as the assessment rules in university admissions or academic title evaluation \citep{harris2022strategic}, and the environments in Generalized Causal Dantzig \citep{long2022generalized}. 

%
Nevertheless, due to data privacy and missing data, interviewers' information is rare in public datasets. Besides, the source label is not always available in some scenarios. For example, people tend to consult an expert consultant, and the consultant's emotional state could be a latent IV that cannot be accessed. 
A large amount of literature for Summary/Selection IVs has attempted to resolve this problem \cite{IVSelection1:burgess2017review,IVSelection4:kuang2020ivy,IVSelection5:hartford2021modeiv,IVSelection6:yuan2022autoiv}.
Two main limitations of these methods are that they require expert knowledge to provide well-predefined IV candidates, and lack metrics to test the validity of IV variables learned by unsupervised methods. 
Moreover, to obtain valid Summary IVs, these methods assume that at least half of pre-defined IV candidates are valid strong IVs so that they can synthesize an IV through a weighted average \cite{IVSelection2:burgess2013use,IVSelection3:davies2015many, IVSelection1:burgess2017review}.

%
Since summary IVs require half of the IV candidates to be valid, which rarely happens in practice, the estimation might be unreliable. 
Therefore, it is highly demanded to model latent IVs and implement a data-driven approach to automatically obtain valid IVs directly from the observed variables $\{X,T,Y\}$, \emph{without pre-defined hand-made IV candidates}. 
Fortunately, the advent of the big data era brought new opportunities to reconstruct IVs from multiple sources data (each source with an independent treatment assignment mechanism).
In the offer case (Fig. \ref{fig1}), the interviewers generate multiple causal relations between the covariates and the treatment, and it can serve as a latent multi-valued IV. \\ \textbf{Motivation}: \emph{Thus, we propose to separate the observational data into multiple groups to reconstruct the source label and then explicitly model the group indicator as a Group Instrumental Variable ({}GIV) to implement IV-based Regression.}

%
In this paper, we aim to recover latent IV and estimate the individual treatment effect (ITE) from mixed observational datasets in the presence of unmeasured confounders.
Due to possibly omitted source labels and unmeasured confounders, traditional methods cannot estimate individual treatment assignment probability and infer treatment effect effectively\footnote{The average treatment assignment mechanism is different from individual mechanisms, which induces additional bias.} 
\cite{ConfMethod8:wu2020dercfr, kuang2017estimating, kuang2020data}. 
Therefore, we propose to reconstruct the source label and model it as a Group Instrumental Variable ({}GIV) to implement IV-based Regression for treatment effect estimation.
In this paper, we conceptualize this line of thought and develop a unified framework (Meta-EM \footnote{Meta means ``learn nonlinear mappings to learn EM''. }) to (1) map the raw data into a representation space to construct Linear Mixed Models for the assigned treatment variable; (2) estimate the distribution differences and model the {}GIV for the different treatment assignment mechanisms using Expectation-Maximization algorithm (EM); and (3) adopt an alternating training strategy to iteratively
optimize the representations and
the joint distribution to model {}GIV for IV regression. Empirical results demonstrate the advantages of the {}GIV compared with SOTA methods.

%
The contribution of our paper is three-fold: 
\begin{itemize}
  \item We propose a Meta-EM algorithm to reconstruct the source label as {}GIV directly from the observed variables, i.e., no available IV candidates for learning, which is beyond the capability of existing Summary IV methods. {}GIV (source label) is effective when there are identifiable differences in mechanisms across groups. 
  \item 
  Meta-EM algorithm uses a shared representation block to learn a nonlinear representation space to EM algorithm, which relaxes the underlying linear regression assumption. Theoretically, Meta-EM can obtain an asymptotic source label as GIV for ITE estimation.
  \item We empirically demonstrate that the Meta-EM algorithm reconstructs the source label as {}GIV from the observed variables for accurate treatment effect estimation and gains SOTA performance compared with existing summary IV methods. 
\end{itemize}

\section{Related Work}
\subsection{Instrumental Variable Methods}
\label{sec:IVMethods}
The sufficient identification results for causal effect under the additive noise assumption in \emph{instrumental variable} regression were developed by \citep{Identify1:angrist1994identification,Identify3:newey2003instrumental,Identify4:hernan2010causal}. For semi-parametric and nonparametric estimation, there are four main research lines about IV methods, including: 
(1) \emph{The two-Stage Least Squares, }Poly2SLS and NN2SLS; (2) \emph{The Kernel-based Methods, } Kernel IV \citep{IVMethod2:singh2019kerneliv} and DualIV \citep{IVMethod3:muandet2019dualiv} map X to a reproducing kernel Hilbert space (RKHS); (3) \emph{The Deep Methods, } DeepIV \citep{IVMethod1:hartford2017deepiv}, OneSIV \citep{IVMethod4:lin2019onesiv} and DFIV \citep{IVMethod7:xu2020dfiv} adopts deep neural nets and fit a mixture density network; (4) \emph{The Adversarial GMM, } AGMM \citep{IVMethod6:lewis2020agmm} and DeepGMM \citep{IVMethod5:2019deepgmm} construct a structural function and select moment conditions via adversarial training.

The above methods are reliable only if the pre-defined IVs are valid and strongly correlated with the treatment. In practice, such valid IVs are hardly satisfied due to the untestable exclusion assumption. In this paper, we reconstruct a {}GIV and plug it into IV methods to predict the treatment effect. 

\subsection{Summary IV Synthesis}
\label{sec:SummaryIV}
A growing number of works have been proposed to synthesize a Summary IV by combining existing IV candidates. In Mendelian Randomization (MR), IV candidates are merged into a summary IV by unweighted/weighted allele scores ({UAS/WAS}) \cite{IVSelection1:burgess2017review,IVSelection4:kuang2020ivy}, \emph{UAS} takes the average of IV candidates while \emph{WAS} weights each IV candidate based on the associations with the treatment. 
Besides, \emph{ModeIV} \cite{IVSelection5:hartford2021modeiv} adopts the tightest cluster center of estimation points as IV to approximate causal effects. Assuming that all IV candidates are independent of the unmeasured confounders, \emph{AutoIV} \cite{IVSelection6:yuan2022autoiv} generates IV representation. 
Existing Summary IV Methods require a high-quality IV candidates' set with at least half valid IVs, which is unrealistic in practice due to cost issues and lack of expert knowledge. Under a more practical setting, we model latent IVs and implement a data-driven approach to automatically reconstruct valid Group IVs directly from the observed variables, without hand-made IV candidates. 

\section{Problem Setup and Assumptions}
\label{sec:assumption}

In this paper, we aim to learn latent IV and estimate the individual treatment effect (ITE) from mixed datasets in the presence of unmeasured confounders. As shown in Fig. \ref{fig1}, a mixed dataset $D=\{ D_1, D_2, \cdots, D_K \}$ collected from $K$ sources 
$D_k= \{x_i, t_i, y_i, \epsilon_i \mid f_{\theta_k}\}_{i=1}^{n_k}, k = 1,2,\cdots,K$, each source with an independent treatment assignment mechanism\footnote{In causal inference, we assume the causal effect of treatment on the outcome is invariant across sources. 
If the treatment effect varies across sources, then we will not identify which treatment effect mechanism the individual's outcome came from in testing.} $f_{\theta_k}$, the size of samples from source $k$ is $n_k$ and the total sample size is $n=\sum_{k=1}^K n_k$.
For unit $i$ from source $z_i=k$, we observe confounders $x_i \in X$ where $X \subset  \mathbb{R}^{m_X}$ with dimension $m_X$, a treatment variable $t_i \in T$ from mechanism $f_{\theta_k}$ where $T \subset  \mathbb{R}$, and a outcome variable $y_i \in Y$ where $Y \subset \mathbb{R}$. In data fusion, due to data privacy and missing data, the source label $z_i$ and some key confounders may be unrecorded in observational data. We incorporate the unobserved confounders into the term $\epsilon_i$. 

Without interactions between unmeasured confounders and treatment, we can represent the effect of infinitely many unmeasured causes as an additive noise $\{\epsilon_{T}, \epsilon_Y\}$ regardless of how they interact among themselves. 
The sufficient identification results for causal effect under \emph{the additive noise assumption} in \emph{instrumental variable} regression were developed by \citep{Identify2:angrist1996identification,Identify3:newey2003instrumental}.

\begin{assumption} \label{ass:noise}
\textbf{Additive Noise Assumption}: Similar to \cite{IVMethod1:hartford2017deepiv,IVMethod2:singh2019kerneliv,IVMethod7:xu2020dfiv}, we assume that the mixed data is generated by: 
\begin{eqnarray}
    \label{eq:Y}
    T = f_{\theta_Z}(X) + \epsilon_{T}, Y = g(T,X) + \epsilon_Y,
\end{eqnarray}
\end{assumption}

\begin{definition}
\textbf{Individual Treatment Effect (ITE)}: \\ $\tau = g(t,X) - g(0,X), {g(t,X) = \mathbb{E}[Y \mid {do}(T=t), X]}$.
\end{definition}


\begin{definition}
\label{assum:IV}
\textbf{An Instrument Variable $Z$} is an exogenous variable that only affects the outcome through its strong association with the treatment. Besides, an valid instrument variable satisfies the following three assumptions:\\
\textbf{Relevance:} interviewers $Z$ assign treatments $T$ to each unit, i.e., $\mathbb{P}(T \mid Z) \neq \mathbb{P}(T)$. \\
\textbf{Exclusion:} interviewers $Z$ does not directly affect the outcome $Y$, i.e., $Z \perp Y \mid T, X,\epsilon$.  \\
\textbf{Unconfounded:} the offer-seekers will be randomly assigned to different interviewers, so $Z$ is independent of all confounders, including $X$ and $\epsilon$, i.e., $Z \perp X,\epsilon$. 
\end{definition}

With the advent of the big data era, a variety of observation databases collected from different sources have been established\nocite{zhang2020devlbert,li2022end}, which may contain the same treatment effect mechanism (from treatment to outcome) but different treatment assignment mechanisms (from covariates to treatment).
Such heterogeneous assignment mechanism is common and widespread in real applications\cite{harris2022strategic, long2022generalized}. 
Plausible settings include: the socio-economic status influences treatment but not outcomes, and admissions assessment rules affect students' SAT scores but do not determine their success in college. In addition, there are many subtle factors that are easily overlooked in real-world applications that may be latent assignment variables, such as weather, holiday, mood, dresses, travel style, lunch, etc. All of them may only affect the treatment choice without directly changing the outcome, but they are often ignored. In the presence of such assignment variables, we propose to separate the observational data into multiple groups to reconstruct the assignment variables and then explicitly model the group indicator as a Group Instrumental Variable ({}GIV) to implement IV-based Regression.

\section{Algorithm} \label{sec:solution}
\begin{figure}[t]
\centering
\includegraphics[width=0.9\columnwidth]{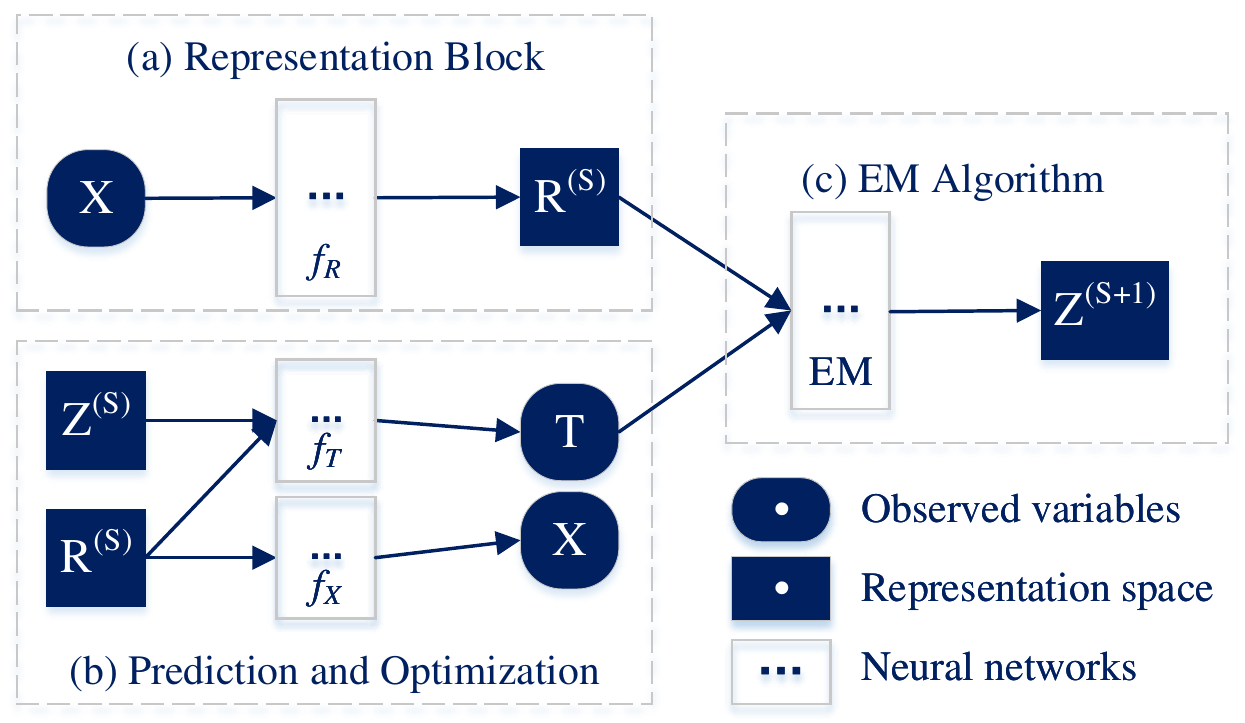} 
\caption{Overview of Meta-EM Architecture.}
\label{fig2}
\end{figure}

In this section, we propose a Meta-EM algorithm to automatically identify the latent source label $Z$, inducing the different treatment assignment mechanisms, as group indicator to separate data into multiple groups. 
Specifically, the overall Meta-EM architecture (Fig. \ref{fig2}) of our model consists of the following components: 
(1) Meta-EM uses a \textbf{Shared Network Layer} to map the covariates $X$ to non-linear representations $R$, and then uses latent variable $Z$ (obtained from EM algorithm) to regress the treatment variables and optimize the representation;
(2) Meta-EM estimates the distribution differences across sources and models latent variable as a GIV for the different treatment assignment mechanisms using \textbf{Expectation-Maximization} algorithm (EM);
(3) Meta-EM adopts an alternating training strategy to iteratively
optimize the \textbf{Representations} and
the joint distribution for \textbf{{}GIV Reconstruction}.
Theoretically, Meta-EM achieve an asymptotic IV and accurately predict ITE by plugging {}GIV into downstream IV-based methods.

\subsection{Representation Learning Step}
Let $z_i = k$ denotes the latent source label ($k=1,2,\cdots,K$) for unit $i$, and source number $K$ will be studied in Sec. \ref{sec:mmd}. 
To construct Linear Mixed Models for the assigned treatment variable $T$, we use a representation function $f_R$ maps the covariates $X \in \mathbb{R}^{m_X}$ into a representation $R \in \mathbb{R}^{m_R}$. 
Consider the following representation model (Fig. \ref{fig2}(a)):
\begin{eqnarray}
    \label{eq:2}
    t_{i} = f_{\theta_{z_i}}(x_i) + \epsilon_{i} = \alpha_{z_i}^{\prime} f_R(x_i) + \epsilon_i = \alpha_{z_i}^{\prime} r_i + \epsilon_i, 
\end{eqnarray}
where  $f_R$ is a shared representation block which can be learned from 
polynomial functions, kernel functions or a neural network, $r_i$ is a (non-)linear representation of $x_i$, and $\alpha_{z_i}$ is the corresponding coefficients for Linear Mixed Models. 
Then we formulate a linear (non-)gaussian mixed model:
\begin{eqnarray}
    \label{eq:3}
    \Scale[1.0]{
    t_{i} = \sum_{k=1}^{K} 1_{z_i=k} \left( \alpha_{k}^{\prime} r_i + \epsilon_i \right)}, 
\end{eqnarray}
where $1_{z_i=k}$ denote the indicator function. Specifically, in polynomial from, we expect to obtain $ t_i =  \Scale[0.96]{\sum_{k=1}^{K} 1_{z_i=k} \left\{\sum_{j=1}^{m_R} \left[ \alpha_{kj}(\xi_{kj,1}x_{ij}^1 + \xi_{kj,2}x_{ij}^2 + \cdots) \right] + \epsilon_i \right\} } $, where $\xi_{kj,d}$ denotes the corresponding expectation coefficient of the $d$-th power of $j$-th variable $x_{ij}$.  

We design two prediction networks $f_T$ and $f_X$ to regress treatment $T$ and covariates $X$, and adopt an alternating training strategy to optimize the representations iteratively (Fig. \ref{fig2}(b)), and the superscript $(s)$ denotes the $s$-th iteration: 
\begin{eqnarray}
    \label{eq:4}
    \nonumber
    \Scale[0.85]{
    \mathcal{L}} & = & \Scale[0.85]{\sum_{i}^{n} \left( f_T(z_i^{(s)},r_i^{(s)}) - t_i \right)^2 + \lambda \sum_{i}^{n} \left( f_X(r_i^{(s)}) - x_i \right)^2} \\
    & = & \Scale[0.85]{\sum_{i}^{n} \left( \alpha_{z_i^{(s)}}^{\prime} r_i^{(s)} - t_i \right)^2 + \lambda \sum_{i}^{n} \left( f_X(r_i^{(s)}) - x_i \right)^2},
\end{eqnarray}
In the term $f_T(z_i^{(s)},r_i^{(s)})$, $z_i^{(s)}$ is a latent function indicator (obtained from Sec. \ref{sec:em}) to activate the corresponding linear coefficients $\alpha_{z_i}$ for treatment regression, and the representation $r_i$ is shared in all sources. The second term is a regularization term to ensure that the representation contains as much information as possible from the original data. 
Besides, $\lambda$ is a trade-off parameter to control the relative importance of treatment regression and covariate regression. We let $\lambda = 1/m_X$, representing that we adopt mean square of $L_2$ norm ($\lambda(f_X(r_i)-x_i)^2=\sum_j(f_j-x_{ij})^2/m_X$, and $j=1,\cdots,m_X$) in covariates regression and treat it as important as treatment regression. 
By minimizing $\mathcal{L}$, our model can map the raw data into a representation space to construct Linear Mixed Models for the assigned treatment variable. 

\subsection{Distribution Learning Step } \label{sec:em}
Based on the traditional Expectation-Maximization (EM) algorithm with group number $K$ (Fig. \ref{fig2}(c)), we seek to find the Maximum Likelihood Estimate (MLE) of the marginal likelihood by iteratively applying the Expectation step and Maximization step. 
Consider the following log-likelihood function for Gaussian mixture:
\begin{eqnarray}
\label{eq:loglikelihood}
    &  & \log {Pr}(C_{TR} \mid \pi, \mu, \Sigma) \nonumber \\
    &= & \Scale[1.0]{\sum_{i=1}^n \sum_{k=1}^K \log \left( \pi_k   {Pr}(c_i \mid \mu_k, \Sigma_k) \right)^{1_{z_i=k}} },
\end{eqnarray}
where $C_{TR}$ denotes the concatenation of $T$ and $R$, $c_i=\{t_i,r_i\}$, and $\pi_k = {Pr}(z_i = k), k=1,2,\cdots,K$. $\mu_k$ and $\Sigma_k$ are the mean vector and covariance matrix of samples $\{c_i\}_{i:z_i=k}$ for group $k$. ${Pr}(c_i \mid \mu_k, \Sigma_k)$ is the density of $c_i$ conditional on $z_i=k$:
\begin{eqnarray}
\Scale[0.9]{{Pr}(c_i \mid \mu_k, \Sigma_k)  =}  \Scale[1.0]{ (2 \pi)^{-\frac{m_R}{2}}\left|\Sigma_{j}\right|^{-\frac{1}{2}} e^{-\frac{1}{2} (c_i-\mu_k)^{\prime} \Sigma_{k}^{-1} (c_i-\mu_k)}}.
\end{eqnarray}

\paragraph{Initialization.}
${Pr}(R^{(s)})$ and ${Pr}(\epsilon)$ should be fixed among all groups since the population does not change according to treatment assignment (Fig. \ref{fig1}), i.e., ${Pr}(R|Z=i)={Pr}(R|Z=j)$ for any groups $Z=i$ and $Z=j$. 
Therefore, we can use $\mathbb{E}[R]$ and $Cov(R,R)$ to initialize the distribution parameters $\theta^{[0]}=\{\pi^{[0]}, \mu^{[0]}, \Sigma^{[0]}\} = \{\{\pi_k^{[0]}, \mu_k^{[0]}, \Sigma_k^{[0]}\}_{k=1,2,\cdots,K}\}$ with $\pi_{k}^{[0]}=1/K$: 
\begin{eqnarray}
\label{eq:init}
    \Scale[0.85]{\mu_k^{[0]} = \{ \mu_k^{[0]}(T), \mathbb{E}[R] \}, \Sigma_k^{[0]} = \left[\begin{array}{cc}
                \sigma_{k}^{[0]}(T,T) & \sigma_{k}^{[0]}(T,R)^T \\
                \sigma_{k}^{[0]}(T,R) & Cov(R,R)
                \end{array}\right]},
\end{eqnarray}
where $\{\mu_k^{[0]}(T), \sigma_{k}^{[0]}(T,T),\sigma_{k}^{[0]}(T,R)\}$ 
are the random initialization of the mean of $T$, the covariance of $T$, and the covariance matrix of $T$ and $R$ in the group $Z=k$, respectively. The superscript $[h]$ denotes $h$-th iteration

\paragraph{Expectation step.}
In the expectation step of the $v$-th iteration, given the observation data $C_{TR}^{(s)}$ and current parameter estimation $\{\pi^{[h]}, \mu^{[h]}, \Sigma^{[h]}\}$, we calculate the log expectation of likelihood function Eq. (\ref{eq:loglikelihood}):
\begin{eqnarray}
\label{eq:expecctation}
    \Scale[0.85]{\mathcal{Q}(\pi^{[h]}, \mu^{[h]}, \Sigma^{[h]}) = 
    \sum_{i=1}^n \sum_{k=1}^K \hat{\gamma}_{ik} \log \left( \pi_k   {Pr}(c_i \mid \mu_k, \Sigma_k) \right)^{1_{z_i=k}} } ,
\end{eqnarray}
where $\hat{\gamma}_{ik}$ is the conditional probability distribution that the $i$-th unit comes from the $k$-th group given $\{\pi^{[h]}, \mu^{[h]}, \Sigma^{[h]}\}$: 
\begin{eqnarray}
\label{eq:ganmaik}
    \Scale[0.9]{\hat{\gamma}_{ik} = {Pr}(z_i=k \mid \pi^{[h]}, \mu^{[h]}, \Sigma^{[h]})}
    = \Scale[1.0]{\frac{\pi_k^{[h]} {Pr}(c_i \mid \mu_k^{[h]}, \Sigma_k^{[h]})}{\sum_{j=1}^K \pi_j^{[h]} {Pr}(c_i \mid \mu_j^{[h]}, \Sigma_j^{[h]})}} .
\end{eqnarray}

\paragraph{Maximization step.} 
In the maximization step of the $h$-th iteration, given the observational data $C_{TR}^{(s)}$ and the current parameter estimation $\theta^{[h]}=\{\pi^{[h]}, \mu^{[h]}, \Sigma^{[h]}\}$, we maximize the expectation of the log-likelihood function $\mathcal{Q}(\{\pi^{[h]}, \mu^{[h]}, \Sigma^{[h]}\})$ (Eq. (\ref{eq:expecctation})) to obtain the parameter estimation $\theta^{[h+1]}$ of next iteration:
\begin{eqnarray}
\label{eq:11}
    \theta^{[h+1]} = \text{argmax}_{\theta} \mathcal{Q}(\{\pi^{[h]}, \mu^{[h]}, \Sigma^{[h]}\}). 
\end{eqnarray}
The solution is: for any $k= 1,2,\cdots,K$, 
\begin{eqnarray}
    \mu_k^{[h+1]} &=& \Scale[1.05]{\frac{\sum_{i=1}^n \hat{\gamma}_{ik} c_i}{\sum_{i=1}^n \hat{\gamma}_{ik}}}, \\
    \Sigma_k^{[h+1]} &=& \Scale[1.05]{\frac{\sum_{i=1}^n \hat{\gamma}_{ik} \left[c_i - \mu_k^{[h+1]} \right]^2}{\sum_{i=1}^n \hat{\gamma}_{ik}}}, \\
    \pi_k^{[h+1]} &=& \Scale[1.05]{\frac{\sum_{i=1}^n \hat{\gamma}_{ik}}{n}}.
\end{eqnarray}

Then, the EM algorithm would obtain the convergent parameters $\theta^*=\{ \pi^*, \mu^*, \Sigma^* \}$ by iteratively applying the Expectation and Maximization steps. We can sample/identify the sub-group indicator $Z^{(s+1)}$ from the estimated distribution parameters $\{ \pi^*, \mu^*, \Sigma^* \}$:
\begin{eqnarray}
\label{eq:reconstruct}
\nonumber
    {\gamma}_{i k}^{(s+1)} = {Pr}(z_i = k) =  \Scale[0.95]{\frac{\pi_k^{*} {Pr}(c_i^{(s)} \mid \mu_k^{*}, \Sigma_k^{*})}{\sum_{j=1}^K \pi_j^{*} {Pr}(c_i^{(s)} \mid \mu_j^{*}, \Sigma_j^{*})}},  \\
    \Scale[0.95]{z_i^{(s+1)} \sim \text{Disc} ({\gamma}_{i 1}^{(s+1)},{\gamma}_{i 2}^{(s+1)},\cdots,{\gamma}_{i K}^{(s+1)})}, 
    \nonumber
\end{eqnarray}
where $\text{Disc}(\cdot)$ denotes discrete distribution with $\{ \gamma_{i k}\}_{k=1}^K$.

\subsection{Optimization and MMD} \label{sec:mmd}
Specifically, our Meta-EM is composed of two phases: in representation learning phase, we fix GIV z to optimize representation network and corresponding coefficients ($f_R$ and $\alpha$) by minimizing objective in Eq. (\ref{eq:4}); in distribution learning phase, we use representation R to re-devide group and update GIV $z$ using EM algorithm.
As shown in Fig \ref{fig2}, the Meta-EM algorithm would learn an optimal representation $R^*$ to learn ${\gamma}_{i k}^{(*)}$ and $ z_i^{(*)}$ by iteratively applying the representation learning step and distribution learning step:
\begin{eqnarray}
\label{eq:reconstruct**}
\nonumber
    {\gamma}_{i k}^{(*)} = {Pr}(z_i = k) =  \Scale[1.0]{\frac{\pi_k^{*} {Pr}(c_i^{(*)} \mid \mu_k^{*}, \Sigma_k^{*})}{\sum_{j=1}^K \pi_j^{*} {Pr}(c_i^{(*)} \mid \mu_j^{*}, \Sigma_j^{*})}}, \\
    z_i^{(*)} \sim \text{Disc} ({\gamma}_{i 1}^{(*)},{\gamma}_{i 2}^{(*)},\cdots,{\gamma}_{i K}^{(*)}).
    \nonumber
\end{eqnarray}
Suppose each coordinate in the coefficient vector $\alpha_k$ in Eq. \eqref{eq:3} is nonzero for all $K=k$. As $(m_R, n) \rightarrow \infty$, $\hat{\gamma}_{i k}$ converges to $1_{z_{i}=k}$ with the rate $o(exp(-(m_R+M)))$ for each $k$, where $m_R$ is the dimension of the representations. 
For theorems and proofs, see the Supplementary material.

As shown in Fig. \ref{fig1}, ${Pr}(X)$ should be fixed among all groups since the population does not change according to treatment assignment, which means the instrumental variable should be independent of all confounders (Unconfounded Assumption of IV), i.e., ${Pr}(X|Z=i)={Pr}(X|Z=j)$ for any groups $Z=i$ and $Z=j$. 
To implement an end-to-end algorithm, we use Maximum Mean Discrepancy (MMD) to measure the correlation between discrete variable $Z$ and observed confounder $X$: $MMD=\| \mathbb{E}[R|Z=i]-\mathbb{E}[X|Z=j] \|^2_2$. Furthermore, we can automatically select the most appropriate group number $K^*$ by the minimum correlation:
\begin{eqnarray}
    \Scale[0.97]{MMD_K = \frac{2}{n(n-1)} \sum_{i=1}^K \sum_{j=i+1}^K ||\bar{X}_{Z=i} - \bar{X}_{Z=j}||_2^2}, \\
    \Scale[0.97]{K^* = \text{argmin}_K MMD_K,K = \{2,3, ...\}}. 
\end{eqnarray}
where $\bar{X}_{Z=i}$ denotes the mean of the covariates $X$ in the $i$-th sub-group according to the EM algorithm.

\section{Experiments}

\subsection{Baselines}
\paragraph{IV generation} In this paper, we adopt Meta-EM with MMD to find the most appropriate group number $K$ and take the cluster results as \textbf{{}GIV$_{EM}$}. We compare our algorithm Meta-EM with the Summary IV methods: (1) \textbf{NoneIV} uses a full-zeros vector as IV; (2) \textbf{UAS} \citep{IVSelection3:davies2015many} takes the average of IV candidates as IV; (3) \textbf{WAS}{ \citep{IVSelection8:burgess2016combining}} weights each candidate based on the associations as IV; (4) \textbf{ModeIV }{\citep{IVSelection5:hartford2021modeiv}} takes the tightest center of estimation points as IV; (5) \textbf{AutoIV}{\citep{IVSelection6:yuan2022autoiv}} learns a disentangled representation as IV. 
Besides, we adopt Meta-KM\footnote{Meta-KM is the K-means replacement of Meta-EM. } to generate \textbf{{}GIV$_{KM}$} and use the superscript $*$ to represent the priori of the number of groups, i.e., \textbf{{}GIV$_{KM}*$}. \textbf{TrueIV} denotes the known ground-truth source label.  

\paragraph{IV regression}
To evaluate the performance of Meta-EM for IV generation, 
we plug synthetic IVs, obtained from Meta-EM and other IV generation baselines, into IV regression methods (as listed in Sec. \ref{sec:IVMethods}) for ITE estimation.

\subsection{Experiments on Synthetic Datasets}
\label{synData}

\begin{figure*}[t]
    \centering
    \includegraphics[width=1.0\textwidth]{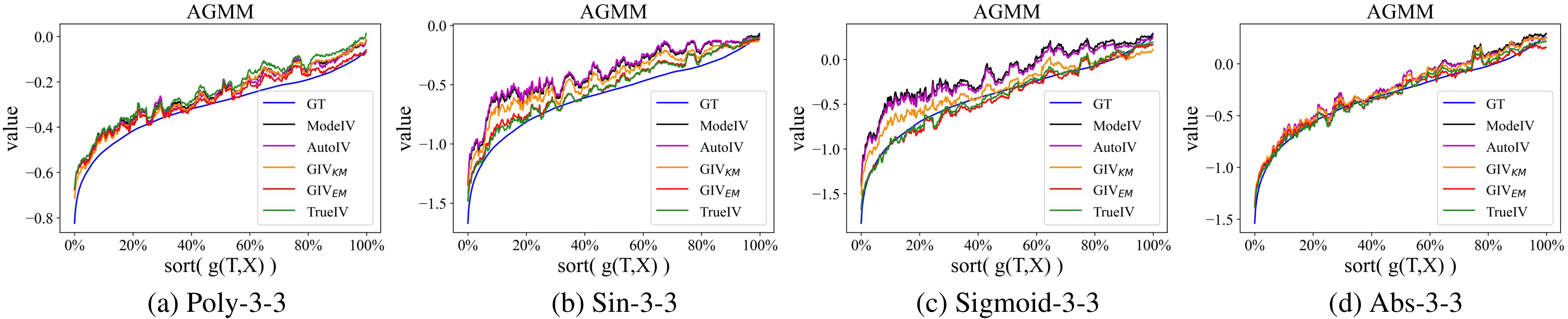}
    \caption{Treatment Effect Estimation (sorted by Ground-Truth $g(T,X)$) in Non-linear Scenario $Data$-$3$-$3$.}
    \label{fig:GIVplot}
\end{figure*}

Similar to DeepIV \cite{IVMethod1:hartford2017deepiv}, DFIV \cite{IVMethod7:xu2020dfiv}, DeepGMM \cite{IVMethod5:2019deepgmm} and AutoIV \cite{IVSelection6:yuan2022autoiv}, due to lack of the prior of latent outcome function and instrumental variable in existing real-world datasets, we evaluate and compare our algorithm Meta-EM with the above baselines on the synthetic and semi-synthetic data.  To simulate real-world data as much as possible, we adjust the difficulty of the simulation and expand experiments to various non-linear scenarios (Fig. \ref{fig:GIVplot}), increase the number of sub-groups and the dimension of covariates (Table \ref{tab:setting} in Supplementary material. The Ablation Experiments for Meta-EM is elaborated in Sec. \ref{sec:ablation}.


\begin{table*}
  \caption{The Mean Squared Error $mean(std)$ on Linear Experiments ($Linear$-3-3)}
  \label{tab:linear}
  \centering
  \scalebox{0.65}{
  \begin{threeparttable}
  \begin{tabular}{ccccccccccc}
    \hline
    & & Poly2SLS &  NN2SLS &  KernelIV &  DualIV$^{(1)}$ &  DeepIV &  OneSIV &  DFIV$^{(1)}$ &  DeepGMM &  AGMM \\
    \hline
     &  NoneIV & 0.330(0.037)  & 1.905(1.255)  & 0.351(0.075)  & 1.924(0.479)  & 0.371(0.011)  & 0.311(0.028)  & 1.335(0.139)  & 0.334(0.068)  & 0.214(0.046) \\
    \hline
    \multirow{4}{*}{\makecell[c]{Summary \\ IV}} &  UAS & 0.330(0.037)  & 2.298(1.461)  & 0.353(0.074)  & \bf 0.976(0.340)  & 0.372(0.024)  & 0.314(0.030)  & 1.299(0.098)  & 0.324(0.037)  & 0.210(0.047) \\
    &  WAS & 0.331(0.038)  & 1.595(0.924)  & 0.360(0.054)  & 2.164(0.459)  & 0.369(0.023)  & 0.339(0.027)  & 1.289(0.123)  & 0.321(0.060)  & 0.234(0.041) \\
    &  ModeIV & 0.330(0.037)  & 2.247(1.300)  & 0.353(0.076)  & 1.899(0.560)  & 0.367(0.021)  & 0.312(0.025)  & 1.292(0.120)  & 0.307(0.070)  & 0.199(0.041) \\
    &  AutoIV & $>100^{(2)}$ & 2.096(1.008)  & 0.352(0.075)  & \bf 0.786(0.324)  & 0.367(0.021)  & 0.311(0.031)  & 1.289(0.113)  & 0.306(0.090)  & 0.214(0.046) \\
    \hline
    \multirow{3}{*}{\makecell[c]{Our \\ Method}} &  {}GIV$_{KM}$ & 0.266(0.128)  & 0.654(0.359)  & 0.222(0.040)  & 1.473(0.259)  & 0.280(0.014)  & 0.230(0.022)  & 1.252(0.110)  & 0.139(0.009)  & 0.123(0.025) \\
    &  {}GIV$_{KM}$* & 0.187(0.087)  & 0.373(0.221)  & 0.214(0.033)  & 1.598(0.361)  & 0.265(0.036)  & 0.219(0.031)  & 1.242(0.097)  & 0.141(0.040)  & 0.100(0.015) \\
    &  {}GIV$_{EM}$ & \bf 0.053(0.004)  & \bf 0.116(0.012)  & \bf 0.112(0.016)  & 1.986(0.414)  & \bf 0.085(0.005)  & \bf 0.116(0.009)  & \bf 0.794(0.079)  & \bf 0.079(0.013)  & \bf 0.064(0.004) \\
    \hline
     &  TrueIV & \bf 0.051(0.004)  & \bf 0.080(0.013)  & \bf 0.108(0.016)  & 1.934(0.389)  & \bf 0.078(0.008)  & \bf 0.112(0.009)  & \bf 0.794(0.065)  & \bf 0.061(0.011)  & \bf 0.062(0.006) \\
  \hline
    \end{tabular}
    \begin{tablenotes}   
        \footnotesize      
        \item[-] (1) DualIV and DFIV don't perform well on {}GIV and even on TrueIV, because they require continuous IVs rather than discrete IVs. (2) "$>100$" means "$\text{MSE}>100$".  
    \end{tablenotes}     
    \end{threeparttable}
    }
\end{table*}


\begin{figure*}[t]
    \centering
    \includegraphics[width=0.85\textwidth]{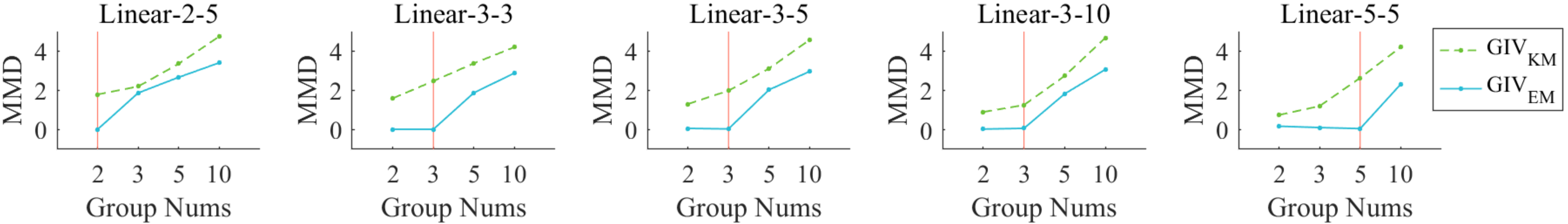}
    \caption{MMD for Selection of Group Number with Different Synthetic Setting ($Data$-$K$-$m_X$). }
    \label{fig:Param}
\end{figure*}


\paragraph{Datasets}
We generate the synthetic datasets as follows:

    \begin{itemize}
        \item \textbf{ The confounders $\{X,\epsilon\}$:}
    \end{itemize}
    \begin{eqnarray}
    \label{eq:confounder}
    \Scale[1.0]{
        X, \epsilon \sim \mathcal{N}(0,\Sigma_{m_X+1}), 
        \Sigma_{m_X+1}=
        \left[\begin{array}{cc}
            \mathrm{I}_{m_{X}} & \sigma \\
            \sigma & 1
        \end{array}\right],
    }
    \end{eqnarray}
    where $m_X$ is the dimensions of observed confounders $X$, $\mathrm{I}_{m_X}$ denotes ${m_X}$ order identity matrix, and $\sigma$ denotes the covariance between confounders $X$ and unmeasured confounder $\epsilon$. In this paper, we let $\sigma = 0.1$. 
    \begin{itemize}
        \item \textbf{ The treatments $T$ collected from multiple sources $Z$:}
    \end{itemize}
    \begin{eqnarray}
    \label{eq:treatment}
        \Scale[0.9]{T = \sum_{z=1}^K 1_{[Z=z]} \left[ \sum_{i=1}^{m_X}w_{zi}[X_i + f_X(X_i)] + f_z(\epsilon) \right] + \delta_T}, \\
        \Scale[0.9]{Z \sim {Pr}(Z=z)= 1 / K, w_{zi}\sim \text{Unif}(-1,1), z=1,\cdots,K}, 
    \label{eq:instrument}
    \end{eqnarray}
    where $X_i,i = \{1,\cdots,m_X\}$ denotes the $i$-th variable in $X$, $\delta_T\sim \mathcal{N}(0,0.1)$, $\text{Unif}$ means we draw $w_{zi}$ from the parameterized uniform distribution, and $f_z(\epsilon)=0.2\epsilon$. The mixed data derives from $K$ different sources, meaning that there are $K$ independent potential treatment assignment models. $Z$ is the indicator of the potential treatment assignment model, which can be regarded as an instrumental variable.
    To simulate real-world data as much as possible, we design 5 different treatment functions $f_X(\cdot)$ to discuss the performance of Meta-EM algorithm: 
    (1) linear scenario, $f_X(X_i)=X_i$; 
    (2) poly scenario, $f_X(X_i)=X_i^2$; 
    (3) sin scenario, $f_X(X_i)=\sin (X_i)$;
    (4) sigmoid scenario, $f_X(X_i)=1 /(1+\exp (- X_i))$;
    (5) abs scenario, $f_X(X_i)=\mathrm{abs}(X_i)$. 

\begin{itemize}
    \item \textbf{The latent outcome function $Y$:}
\end{itemize}
    \begin{eqnarray}
    \label{eq:outcome}
        \nonumber
        \Scale[0.95]{Y} = & - & \Scale[0.95]{1.5T+0.9T^2 + \sum_{i=1}^{m} \frac{X_i}{m}+|X_1 X_2|} \\ 
        & - & \Scale[0.95]{\text{sin}(10+X_2X_3) + 2\epsilon + \delta_Y}. 
    \end{eqnarray}
    where $\epsilon$ is an unmeasured confounder and $\delta_Y \sim \mathcal{N}(0,0.1)$.


\begin{table}
  \caption{Ablation Experiments for Meta-EM in $Data$-$2$-$5$}
  \centering
  \label{tab:ablation}
  \scalebox{0.75}{
  \begin{tabular}{c ccccc}
    \hline
    & Linear & Poly & Sin & Sigmoid & Abs \\
    \hline
    EM & 86.1\% & 82.9\% & 87.2\% & 89.4\% & 85.2\% \\
    Meta-EM & \bf 86.1\% & \bf 92.7\% & \bf 90.3\% & \bf 94.7\% & \bf 92.3\% \\
    Fun-EM & \bf 86.1\% & \bf 98.1\% & \bf 96.3\% & \bf 98.4\% & \bf 96.5\% \\
  \hline
  \end{tabular}
  }
\end{table}


For synthetic datasets, we sample 3,000 units and perform 10 independent replications to report mean squared error (MSE) and standard deviations of the individual treatment effect estimation over the testing data (3000 units) that we intervene the treatment as $T=do(t)$. To verify the effectiveness of {}GIV$_{EM}$ in different scenarios with different dimensions of covariates $m_X$ and different group numbers $K$, we use \textbf{$Data$-$K$-$m_X$} to denote the different scenarios. In this paper, we set the representation dimension as $m_R = m_X$. 


\begin{table*}
  \caption{The Mean Squared Error $mean(std)$ on IHDP \& PM-CMR Dataset}
  \centering
  \label{tab:True2-1}
  \scalebox{0.65}{
  \begin{tabular}{cc cccc| cccc}
    \hline
    & & \multicolumn{4}{c}{IHDP Dataset} & \multicolumn{4}{c}{PM-CMR Dataset} \\
    \cmidrule(r){3-6} \cmidrule(r){7-10}
    & & Poly2SLS &  KernelIV &  DeepIV &  AGMM &  Poly2SLS &  KernelIV &  DeepIV &  AGMM \\
    \hline
    & NoneIV & 0.238(0.132)  & 0.456(0.243)  & 0.583(0.240)  & 0.140(0.063) & 0.181(0.044)  & 0.352(0.198)  & 0.409(0.160)  & 0.130(0.064) \\
    \hline
    \multirow{4}{*}{\makecell[c]{Summary \\ IV}} & UAS & 0.238(0.133)  & 0.457(0.244)  & 0.574(0.245)  & 0.142(0.061) & 0.181(0.044)  & 0.352(0.198)  & 0.404(0.164)  & 0.128(0.064) \\
    & WAS & 0.239(0.134)  & 0.455(0.241)  & 0.567(0.226)  & 0.144(0.059) & 0.181(0.044)  & 0.372(0.207)  & 0.417(0.164)  & 0.157(0.080) \\
    & ModeIV & 0.240(0.133)  & 0.460(0.246)  & 0.572(0.245)  & 0.149(0.060) & 0.181(0.044)  & 0.359(0.201)  & 0.406(0.150)  & 0.131(0.070) \\
    & AutoIV & $>100$ & 0.457(0.243)  & 0.583(0.250)  & 0.142(0.069) & 0.179(0.044)  & 0.351(0.198)  & 0.409(0.180)  & 0.129(0.064) \\
    \hline
    \multirow{2}{*}{\makecell[c]{Our Method}} & {}GIV$_{KM}$ & 0.078(0.029)  & 0.354(0.179)  & 0.505(0.203)  & 0.112(0.050) & 0.088(0.044)  & 0.329(0.202)  & 0.381(0.165)  & 0.117(0.053) \\
    & {}GIV$_{EM}$ & \bf 0.034(0.011)  & \bf 0.202(0.173)  & \bf 0.482(0.228)  & \bf 0.095(0.035) & \bf 0.048(0.012)  & \bf 0.308(0.210)  & \bf 0.339(0.184)  & \bf 0.085(0.045) \\
    \hline
    & TrueIV & \bf 0.033(0.009)  & \bf 0.151(0.060)  & \bf 0.458(0.166)  & \bf 0.093(0.035) & \bf 0.028(0.007)  & \bf 0.140(0.074)  & \bf 0.141(0.054)  & \bf 0.054(0.023) \\
  \hline
  \end{tabular}
  }
\end{table*}


\begin{figure}
    \centering
    \includegraphics[width=0.45\textwidth]{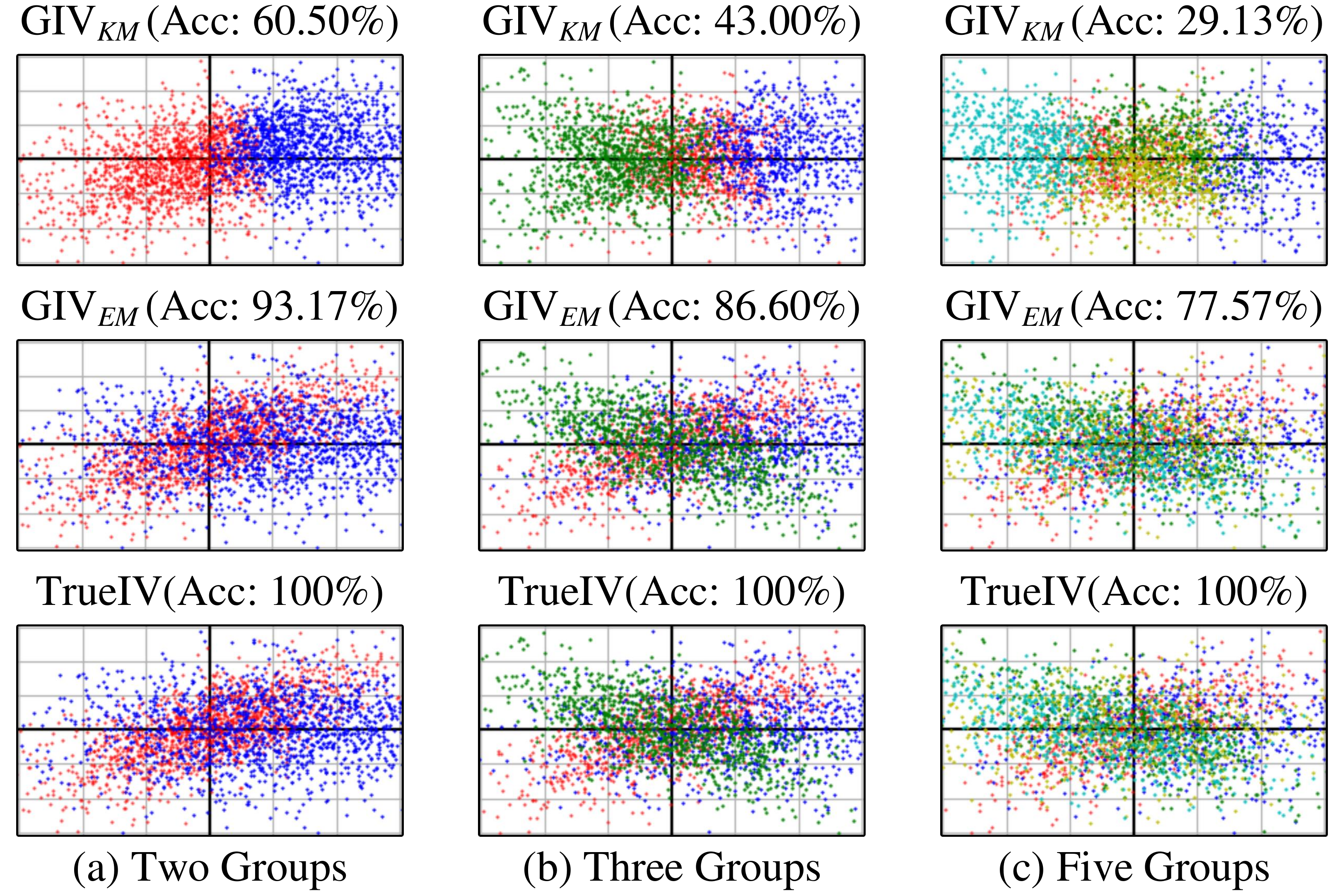}
    \caption{Reconstruction Accuracy of Meta-KM \& Meta-EM with Different Group Number.}
    \label{fig:GIVcluster}
\end{figure}


\paragraph{The Results of Individual Treatment Effect Estimation}
As shown in Table \ref{tab:linear} (The top-2 performance is highlighted in bold for all tables), we compare the performance of {}GIV with other Summary IVs and TureIV on various downstream IV methods in linear scenario ($Linear$-3-3) with $T=do(t)$. 
Following observations are identified from the results: 
(1) Without valid IV candidates, Summary IVs are not reliable and fail to synthesize a valid IV, and plugging them into the IV methods can hardly improve the estimation performance, which is close to the NoneIV; 
(2) DualIV and DFIV do not perform well on {}GIV and fail to estimate treatment effect, even with TrueIV, because they require continuous IVs rather than discrete IVs. 
(3) Through clustering, we reconstruct the latent exogenous IV that generates different treatment mechanisms, {}GIVs (with Meta-KM or Meta-EM) bring higher accuracy on individual treatment effect estimation by comparing with Summary IV methods in various IV-based methods except for DualIV; 
(4) By estimating the latent differentiated covariate-treatment distribution parameters across groups and reconstructing the source label, {}GIV$_{EM}$ significantly improves the performance of clustering methods compared with {}GIV$_{KM}$ and achieves SOTA performance for individual treatment effect estimation, even comparable with TrueIV. Empirically, this demonstrates that our Meta-EM successfully reconstructs the GIV, and it converges to the TrueIV, i.e., source label. 

Then, to verify the effectiveness of {}GIV in non-linear cases, we design 4 different non-linear treatment functions $f_X(\cdot)$ to evaluate the treatment effect estimation performance of Meta-EM algorithm. We select the SOTA IV-based method (AGMM) to evaluate {}GIV. We plot the estimated value of effect function with T=do(t) and sort it by Ground-Truth (GT) for different synthetic scenarios. The results (Fig. \ref{fig:GIVplot}) show {}GIVs (with Meta-KM or Meta-EM) achieve SOTA performance, especially {}GIV$_{EM}$ achieves comparable results with TrueIV and estimated outcome curves from {}GIV$_{EM}$ approximate the true curve. 
For the detailed results of non-linear cases, see the Supplementary material.

\paragraph{The Ablation Study for Reconstruction Accuracy of GIV}
\label{sec:ablation}
To demonstrate that Meta-EM can automatically find the proper group number and implement end-to-end train for IV generation, we plot MMD line for each group number in different synthetic settings (Linear-$K$-$m_X$). As shown in Fig. \ref{fig:Param}, Meta-EM always find the proper group number (red-line) automatically, but Meta-KM fails to do it. Besides, as an ablation experiment, we compare the accuracy of Meta-KM and Meta-EM for GIV reconstruction on data fusion with different group numbers. 
As shown in Fig. \ref{fig:GIVcluster}, Meta-EM algorithm successfully reconstructs the GIV, and the average reconstruction accuracy has reached 77\% under various group numbers, especially exceeding 90\% accuracy on Two Groups setting. In contrast, the identification accuracy of Meta-KM is basically below 60\%. 

Meta-EM algorithm uses a shared representation block to learn a nonlinear representation space to EM algorithm, which relaxes the underlying linear regression assumption. To verify it, in the ablation experiments (Table \ref{tab:ablation}), we compare the accuracy of EM, Meta-EM and Fun-EM, where Fun-EM implements EM algorithm with known non-linear functions $f_X(X)$ (Eq. (\ref{eq:treatment})). The results show that Meta-EM improves the reconstruction accuracy by 6.3\% than EM algorithm, bus still below the GT Fun-EM.

\subsection{Experiments on Real-world Datasets}
\label{semiData}

\paragraph{Real-world datasets} \label{sec:realdata1}
Similar to previous methods\citep{ConfMethod10:nie2020vcnet,IVMethod1:hartford2017deepiv,ConfMethod11:bica2020scigan,ConfMethod9:schwab2020drnet}, we perform experiments on two semi-synthetic real-world datasets \textbf{IHDP}\footnote{IHDP: https://www.fredjo.com/}  \citep{ConfMethod2:shalit2017cfr} \& \textbf{PM-CMR}\footnote{PM-CMR:https://pasteur.epa.gov/uploads/10.23719/1506014/\\SES\_PM25\_CMR\_data.zip} \citep{ConfMethod13:wyatt2020dataset3}, as the true effect function (the counterfactual outcomes) is rarely available for real-world data. Then we use the continuous variables from \textbf{IHDP} \& \textbf{PM-CMR} to replace the covariates $X$ in Eq. (\ref{eq:treatment})\&(\ref{eq:outcome}) to generate treatment $T$ and outcome $Y$, respectively. 
Both two datasets are randomly split into training (63\%), validation (27\%), and testing (10\%). We perform 10 replications to report the mean squared error (MSE) and its standard deviations (std) of the treatment effect function estimation. We select four SOTA IV-based methods to evaluate the performance of GIV. For the detailed description and the full results about real-world datasets, see the Supplementary material.

\paragraph{The results of individual treatment effect estimation} 
By estimating the latent differentiated covariate-treatment distribution parameters across groups, Meta-EM reconstructs the latent IV and the reconstruction accuracy reaches 93.47\% and 82.62\% on \textbf{IHDP} and \textbf{PM-CMR}, however, K-Means is only 64.29\% and 46.09\%. This demonstrates Meta-EM can automatically find the optimal IV, but K-Means cannot. 
In Table \ref{tab:True2-1}, comparing the two optimal combinations (AutoIV in Poly2SLS \& UAS in AGMM) in effect estimation in Table \ref{tab:True2-1}, Meta-EM further reduced the errors by 0.131(↓73\%) and 0.043(↓33\%), which well eliminated the unmeasured confounding bias.
Besides, GIV$_{EM}$ shows consistent and robust performance, always maintaining the performance of top-2 and almost achieving the same effect as TrueIV on IHDP \& PM-CMR Datasets. Compared with {}GIV$_{EM}$, the performance of {}GIV$_{KM}$ exceeds most baselines in downstream tasks, but it is still inferior to {}GIV$_{EM}$ and TrueIV. 
This means that GIV$_{EM}$ can reconstruct the latent group IV with the data distribution in the real scene and obtain asymptotically unbiased causal effect estimation.

\section{Conclusion}
The advent of the big data era brought new opportunities and challenges to draw treatment effect. In the causal effect estimation, unmeasured confounders would fool the estimator to draw erroneous conclusions, whereas traditional IV solutions to this problem rely on valid IV candidates from expert knowledge or RCTs. To relax this requirement, by estimating the differentiated covariate-treatment distribution across groups, we propose a novel Meta-EM, a tool for reconstructing latent Group IVs and predicting treatment effect function from data fusion. 
To the best of our knowledge, using representation learning as a non-linear bridge to reconstruct Group Instrumental Variable by Meta-EM algorithm in data fusion is the first work for IV generation without expert knowledge. Theoretically and empirically, we address a vital problem in causal inference: how to reconstruct valid IVs from data fusion for ITE without any prior knowledge.

\newpage
\balance
\bibliography{ref}

\clearpage
\appendix

\section{The Convergence Analysis of Meta-EM}
\label{sec:convergence}

\begin{theorem} 
\textbf{Distribution Learning Step}. Meta-EM learns a non-linear representation to construct Linear Mixed Models for the assigned treatment variable. Based on the representation, we model the likelihood function for observational data $c_i = \{t_i,r_i\}, i = 1,2,\cdots,n$ as ${Pr}(C_{TR} \mid \theta=\{\pi, \mu, \Sigma\})$.
We denote parameter sequence estimated by EM algorithm as $\theta^{[h]} (h=1,2,\cdots)$, and denote the corresponding likelihood function sequence as ${Pr}(C_{TR} \mid \theta^{[h]}) (h=1,2,\cdots)$. Then ${Pr}(C_{TR} \mid \theta^{[h]})$ is a monotonic sequence which constantly increase:
\begin{eqnarray}
{Pr}(C_{TR} \mid \theta^{[h+1]}) \geq {Pr}(C_{TR} \mid \theta^{[h]}).
\end{eqnarray}
\end{theorem}

\paragraph{Proof}
The observation:
\begin{eqnarray}
\nonumber
\Scale[0.95]{{Pr}(C_{TR} \mid \theta) = \frac{{Pr}(C_{TR}, Z \mid \theta)}{{Pr}(Z\mid C_{TR}, \theta)}},
\end{eqnarray}
\begin{eqnarray}
\Scale[0.95]{\log {Pr}(C_{TR} \mid \theta) = \log {{Pr}(C_{TR}, Z \mid \theta)} - \log {{Pr}(Z\mid C_{TR}, \theta)}}.
\nonumber
\end{eqnarray}

Take the expectation of the log likelihood function $\mathcal{Q}(\theta, \theta^{[h]})$:
\begin{eqnarray}
\mathcal{Q}(\theta, \theta^{[h]}) = \Sigma_Z \log {{Pr}(C_{TR}, Z \mid \theta)} {Pr}(Z \mid C_{TR}, \theta^{[h]}), 
\end{eqnarray}
Let
\begin{eqnarray}
\mathcal{H}(\theta, \theta^{[h]}) = \Sigma_Z \log {{Pr}(Z \mid C_{TR}, \theta)} {Pr}(Z\mid C_{TR}, \theta^{[h]}), 
\end{eqnarray}
Then,
\begin{eqnarray}
\nonumber \log {Pr}(C_{TR} \mid \theta^{[h+1]}) - \log {Pr}(C_{TR} \mid \theta^{[h]}) \\
\nonumber \Scale[0.9]{ = [\mathcal{Q}(\theta^{[h+1]}, \theta^{[h]}) - \mathcal{H}(\theta^{[h+1]}, \theta^{[h]})] 
- [\mathcal{Q}(\theta^{[h]}, \theta^{[h]}) - \mathcal{H}(\theta^{[h]}, \theta^{[h]})]} \\
\nonumber = \Scale[0.9]{[\mathcal{Q}(\theta^{[h+1]}, \theta^{[h]}) - \mathcal{Q}(\theta^{[h]}, \theta^{[h]})] 
- [\mathcal{H}(\theta^{[h+1]}, \theta^{[h]}) - \mathcal{H}(\theta^{[h]}, \theta^{[h]})]}
\end{eqnarray}
where, the term $[\mathcal{Q}(\theta^{[h+1]}, \theta^{[h]}) - \mathcal{Q}(\theta^{[h]}, \theta^{[h]})] \geq 0$, because we maximize the expectation of the log likelihood function to obtain the parameter estimation $\theta^{[h+1]}$ of next iteration:
\begin{eqnarray}
    \theta^{[h+1]} = \text{argmax}_{\theta} \mathcal{Q}(\theta, \theta^{[h]}). 
\end{eqnarray}

The key component of EM algorithm is the use of Jensen's inequality:
\begin{eqnarray}
    \nonumber &   & \mathcal{H}(\theta^{[h+1]}, \theta^{[h]}) - \mathcal{H}(\theta^{[h]}, \theta^{[h]}) \\
    \nonumber & = & \Sigma_Z \left( \log \frac{{Pr}(Z\mid C_{TR}, \theta^{[h+1]})}{{Pr}(Z\mid C_{TR}, \theta^{[h]})} \right)  {Pr}(Z\mid C_{TR}, \theta^{[h]}) \\
    \nonumber & \leq & \log \left( \Sigma_Z \frac{{Pr}(Z\mid C_{TR}, \theta^{[h+1]})}{{Pr}(Z\mid C_{TR}, \theta^{[h]})} {Pr}(Z\mid C_{TR}, \theta^{[h]}) \right) \\
    & = & \log \left( \Sigma_Z {Pr}(Z\mid C_{TR}, \theta^{[h+1]})\right) = 0.
\end{eqnarray}

In conclusion, we obtain $ \log {Pr}(C_{TR} \mid \theta^{[h+1]}) - \log {Pr}(C_{TR} \mid \theta^{[h]}) \geq 0$, that means ${Pr}(C_{TR} \mid \theta^{[h]})$ is a monotonic sequence which constantly increase:
\begin{eqnarray}
{Pr}(C_{TR} \mid \theta^{[h+1]}) \geq {Pr}(C_{TR} \mid \theta^{[h]}).
\end{eqnarray}
If there is an upper bound for ${Pr}(C_{TR} \mid \theta)$, the sequence $\log {Pr}(C_{TR} \mid \theta^{[h]}) (i=1,2,\cdots)$ would converge to a specific value $L^*$.

\begin{theorem} 
\textbf{Representation Learning Step}. 
In $s$-th iteration, Meta-EM learns a latent source variable as a group instrumental variable ({}GIV) $Z^{(s)}$ indicating multiple treatment assignment mechanisms. Based on {}GIV $Z^{(s)}$, We use multiple linear functions indicated by $Z^{(s)}$ to model Linear Mixed Models explicitly to optimize the shared representation $R^{(s)}$. We denote {}GIV sequence and representation sequence as $Z^{(s)}$ and $R^{(s)} (s=1,2,\cdots)$, separately, and let $C_{TR}^{(s)}$ as the concatenation of $T$ and $R^{(s)}$. The corresponding likelihood function sequence as ${Pr}(Z^{(s+1)} \mid C_{TR}^{(s)})$. Then, ${Pr}(Z^{(s+1)} \mid C_{TR}^{(s)})$ is a monotonic sequence which constantly increase:
\begin{eqnarray}
{Pr}(Z \mid C_{TR}^{(s+1)}) \geq {Pr}(Z \mid C_{TR}^{(s)}).
\end{eqnarray}
\end{theorem}

\paragraph{Proof} 
In the representation learning step of $s$-th iteration, the Meta-EM algorithm uses a shared representation block to learn a non-linear representation $R=f_R(X), R \in \mathcal{R}^{m_R}$ to regress treatment using Linear Mixed Models indicated by {}GIV $Z$. By minimizing the regression error, the representation module will capture the non-linear terms of the raw data on the treatment variables. As missing domain labels, the {}GIV $Z$ will bring additional information for representation learning. Therefore, the higher the reconstruction accuracy of the instrumental variable $Z$, the more accurate the non-linear terms from representation learning will be. Further, the performance of the EM algorithm will also be improved by using these non-linear representations to construct linear mixed models of the treatment variables, i.e., ${Pr}(Z \mid C_{TR}^{(s+1)}) \geq {Pr}(Z \mid C_{TR}^{(s)})$. It is a Mutual Reinforcement Learning process: learn representation to learn IV, and then learn IV to learn representation at each iteration. 

Besides, to ensure that the representation learning does not lose information from the raw data, we also construct a covariate reconstruction loss function and minimize the term. The objective is to minimize $\mathcal{L} = \sum_{i}^{n} \left( f_T(z_i^{(s)},r_i^{(s)}) - t_i \right)^2 + \lambda \sum_{i}^{n} \left( f_X(r_i^{(s)}) - x_i \right)^2$. 

\paragraph{Remark} Theoretically, the Meta-EM algorithm is effective when identifiable differences in treatment assignment mechanisms across groups exist. Overall, the reconstruction accuracy has reached 77\% in Section \ref{sec:ablation}, and we can estimate the treatment effect function accurately. 

\section{Proof of Theorem} \label{sec:proof}

\subsection{Asymptotic Results of GIV}
In Meta-EM, $\hat{\gamma}_{i k}$ converges to $1_{z_{i}=k}$ with the rate $o(exp(-(m_R+M)))$, where $m_R$ is the dimension of the representations. This rate comes from bounding the probability of being in the wrong group, which can be shown by using that the density function of each group in EM algorithm follows a normal distribution, and the tail of the normal distribution is exponentially bounded and linear in $m_R$.  Note that this result does not directly involve the sample size n. This is because the estimation error of the representations is not the leading term in the estimation error of $\hat{\gamma}_{i k}$. However, we note that as $n$ gets larger, the representations can be learned more precisely, which may have small effects on $\hat{\gamma}_{i k}$. 
This implies that the Meta-EM algorithm can asymptotically recover the true group label for each unit $i$ in large samples. 
\begin{theorem} \textbf{Asymptotic Results of GIV. } \\
Suppose each coordinate in the coefficient vector $\alpha_k$ in Eq. \eqref{eq:3} is nonzero for all $k$.
As $(m_R, n) \rightarrow \infty$, for each $k$:\\
(1) $\hat{\gamma}_{i k} \xrightarrow{p} 1_{z_{i}=k}$, \\
(2) $\hat{z}_i \sim \text{Disc} ({\hat \gamma}_{i 1},{\hat \gamma}_{i 2},\cdots,{\hat \gamma}_{i K})$ is an asymptotic IV, i.e., $Pr(\hat{Z}, X, T, Y) \xrightarrow{p} Pr(Z, X, T, Y)$. \\
where, $\xrightarrow{p}$ denotes convergence in probability.
\end{theorem}


\paragraph{Proof}
\textbf{(1)} Consider the following representation model (Eq. \eqref{eq:3}):
\begin{eqnarray*}
\label{eq:proof1}
t_i = \alpha_k^{\prime} r_i + \epsilon_i, \text{ if } z_i = k, 
\end{eqnarray*}
where $\alpha_k$ is a $m_R$ dimensional vector of coefficients for source $k$, and each coordinate in the coefficient vector $\alpha_k$ is nonzero for all $k$. $r_i$ is a $m_R$ dimensional vector of representation for unit $x_i$,
$z_{t}$ is the source label indicating which unit $x_{i}$ belongs to, and $e_{i}$ is the error term allowed to have cross-sectional and heteroskedasticity. $\{x_{i}, t_{i}\}$ is observable and $\{\alpha_k, r_i, z_i, \epsilon_i\}$ are unobservable.

Let $c_i$ as the concatenation of $t_i$ and $r_i$, i.e., $c_i = (t_i, r_i)$:
\begin{eqnarray*}
c_i = (\alpha_k,I)^{\prime} r_i + \epsilon_i, \text{ if } z_i = k, 
\end{eqnarray*}
where $I=\{e_j\}_{j=1}^{m_R}$ is an identity matrix of size $m_R$, and $e_{j}=(0, \ldots, 0, \underset{j-th}{1}, 0, \ldots, 0)^\prime$: 
\begin{eqnarray*}
c_i = A_k r_i + \epsilon_i = (\alpha_k,e_1,e_2,\cdots,e_{m_R})^{\prime} r_i + \epsilon_i, \text{ if } z_i = k.
\end{eqnarray*}
Besides,
\begin{eqnarray*}
    {Pr}(c_i \mid \mu_k, \Sigma_k) =(2 \pi)^{-\frac{m_R}{2}}\left|\Sigma_{k}\right|^{-\frac{1}{2}} e^{-\frac{1}{2} (c_i-\mu_k)^{\prime} \Sigma_{k}^{-1} (c_i-\mu_k)}
\end{eqnarray*}

From Eq. (\ref{eq:ganmaik}), we have:
\begin{eqnarray*}
    \hat{\gamma}_{ik} 
    = \frac{\pi_k {Pr}(c_i \mid \hat{\mu}_k, \hat{\Sigma}_k)}{\sum_{j=1}^K \pi_j {Pr}(c_i \mid \hat{\mu}_j,
    \text{and} \hat{\Sigma}_j)}, \sum_{k=1}^K \hat{\gamma}_{ik} = 1 .
\end{eqnarray*}
From the Convergence Analysis of Meta-EM:
\begin{eqnarray*}
    \hat{\mu}_k = \frac{\sum_{i=1}^n \hat{\gamma}_{ik} c_i}{\sum_{i=1}^n \hat{\gamma}_{ik}}, 
    \hat{\Sigma}_k = \frac{\sum_{i=1}^n \hat{\gamma}_{ik} \left[c_i - \hat{\mu}_k \right]^2}{\sum_{i=1}^n \hat{\gamma}_{ik}}
\end{eqnarray*}
If $n \rightarrow \infty$, then the number of samples for each group $n_k = n \times Pr(Z=k) \rightarrow \infty$ and we have $\hat{\mu}_{k} \xrightarrow{p} {\mu}_{k}$ and $\hat{\Sigma}_{k} \xrightarrow{p} {\Sigma}_{k}$ for each $k=1,2,\cdots,K$.

If the dimension of representations is sufficiently large and the distribution of representations for different groups can be separated with high probability, i.e., $Pr(z_{i}=k) \xrightarrow{p} 1_{z_{i}=k}$, then for each unit $z_i=k$ as $(m_R, n) \rightarrow \infty$:
\begin{eqnarray}
\nonumber
\left|\hat{\gamma}_{i k}-1_{z_{i}=k}\right| = \frac{
\sum_{j \not = k} \pi_j {Pr}(c_i \mid \hat{\mu}_j, \hat{\Sigma}_j)
}{
\sum_{j=1}^K \pi_j {Pr}(c_i \mid \hat{\mu}_j, \hat{\Sigma}_j)
} \\
\leq \sum_{j \not = k} \frac{\pi_j}{\pi_k} \exp\{\log \frac{ {Pr}(c_i \mid \hat{\mu}_j, \hat{\Sigma}_j)}{ {Pr}(c_i \mid \hat{\mu}_k, \hat{\Sigma}_k)}\}
\end{eqnarray}

If $z_i = v \not = k$, then:
\begin{eqnarray}
\nonumber
\left|\hat{\gamma}_{i k}-1_{z_{i}=k}\right| = \frac{
\pi_k {Pr}(c_i \mid \hat{\mu}_k, \hat{\Sigma}_k)
}{
\sum_{j=1}^K \pi_j {Pr}(c_i \mid \hat{\mu}_j, \hat{\Sigma}_j)
} \\
\leq \frac{\pi_k}{\pi_v} \exp\{\log \frac{ {Pr}(c_i \mid \hat{\mu}_k, \hat{\Sigma}_k)}{ {Pr}(c_i \mid \hat{\mu}_v, \hat{\Sigma}_v)}\}
\end{eqnarray}

Consider $z_i=k$, then for any $j \not= k$ and for a sufficiently large $M > 0$:
\begin{eqnarray*}
{Pr}\left( 
\sup_i \left[ 
\log \frac{ {Pr}(c_i \mid \hat{\mu}_j, \hat{\Sigma}_j)}{ {Pr}(c_i \mid \hat{\mu}_k, \hat{\Sigma}_k)}
\right] \geq - (m_R + M)
\right) \rightarrow 0, 
\end{eqnarray*}
That is, 
\begin{eqnarray*}
{Pr}\left( 
\min_i \left[ 
\log \frac{ {Pr}(c_i \mid \hat{\mu}_k, \hat{\Sigma}_k)}{ {Pr}(c_i \mid \hat{\mu}_j, \hat{\Sigma}_j)}
\right] \leq m_R + M
\right) \rightarrow 0,
\end{eqnarray*}

Then, 
\begin{eqnarray*}
& &    \log \frac{ {Pr}(c_i \mid \hat{\mu}_k, \hat{\Sigma}_k)}{ {Pr}(c_i \mid \hat{\mu}_j, \hat{\Sigma}_j)} \\
& = & \log \frac{ \left|\hat{\Sigma}_{k}\right|^{-\frac{1}{2}} e^{-\frac{1}{2} (c_i-\hat{\mu}_k)^{\prime} \hat{\Sigma}_{k}^{-1} (c_i-\hat{\mu}_k)}}
{ \left|\hat{\Sigma}_{j}\right|^{-\frac{1}{2}} e^{-\frac{1}{2} (c_i-\hat{\mu}_j)^{\prime} \hat{\Sigma}_{j}^{-1} (c_i-\hat{\mu}_j)}} \\
& = & -\frac{1}{2} \log \frac{\left|\hat{\Sigma}_{k}\right|}{\left|\hat{\Sigma}_{j}\right|} -\frac{1}{2} \log \frac{\exp (c_i-\hat{\mu}_k)^{\prime} \hat{\Sigma}_{k}^{-1} (c_i-\hat{\mu}_k)}{\exp (c_i-\hat{\mu}_j)^{\prime} \hat{\Sigma}_{j}^{-1} (c_i-\hat{\mu}_j)}
\end{eqnarray*}

Let $B_k=\bar{c}_i^{[k]} +\epsilon_i= c_i-\hat{\mu}_k$:
\begin{eqnarray}
\bar{c}_i^{[k]} +\epsilon_i= A_k r_i - 
A_k \frac{\sum_{i=1}^{n} \hat{\gamma}_{i k} r_i}{\sum_{i=1}^{n} \hat{\gamma}_{i k}} = A_k (r_i-\bar{r}_i[k]) +\epsilon_i
\end{eqnarray}
where $\bar{r}_i[k] = \frac{\sum_{i=1}^{n} \hat{\gamma}_{i k} r_i}{\sum_{i=1}^{n} \hat{\gamma}_{i k}}$ denotes the mean of $r_i$ on source $k$. 
Then, 
\begin{eqnarray}
\hat{\Sigma}_{k} =  \bar{C}[k]\bar{C}[k]^\prime + \sigma^2I_{m_R+1}
\end{eqnarray}
where $\bar{C}[k]^{-1} = (\bar{c}_{i1}^{[k]},\bar{c}_{i2}^{[k]},\cdots,\bar{c}_{i(m_R+1)}^{[k]})^\prime$, and $\epsilon \sim \mathcal{N}(0, \sigma^2)$ from additive noise assumption.

\noindent By Woodbury matrix identity, 
\begin{eqnarray}
\hat{\Sigma}_{k}^{-1} = \sigma^{-2}I - \sigma^{-2} \bar{C}[k] (\sigma^2I + \bar{C}[k]^\prime\bar{C}[k])^{-1} \bar{C}[k]^\prime
\end{eqnarray}
and
\begin{eqnarray}
\nonumber
(\sigma^2I + \bar{C}[k]^\prime\bar{C}[k])^{-1} = (\bar{C}[k]^\prime\bar{C}[k])^{-1} \\
+ \sigma^2  
(\sigma^2I + \bar{C}[k]^\prime\bar{C}[k])^{-1}
(\bar{C}[k]^\prime\bar{C}[k])^{-1}
\end{eqnarray}
We have,
\begin{eqnarray*}
& & \Scale[0.9]{\left(\bar{c}_i^{[k]} +\epsilon_i\right)^{\prime} \hat{\Sigma}_{k}^{-1}\left(\bar{c}_i^{[k]} +\epsilon_i\right) 
= B_k^{\prime}
\hat{\Sigma}_{k}^{-1}
B_k }\\
&=& \Scale[0.9]{\sigma^{-2} B_k^{\prime} (I-\bar{C}[k] (\bar{C}[k]^\prime\bar{C}[k])^{-1}\bar{C}[k]^\prime)B_k }\\
&- &\Scale[0.9]{ B_k^{\prime}
\bar{C}[k](\sigma^2I + \bar{C}[k]^\prime\bar{C}[k])^{-1}(\bar{C}[k]^\prime\bar{C}[k])^{-1}\bar{C}[k]^\prime
B_k}
\end{eqnarray*}
Let $M_1[k] = \bar{C}[k]^\prime\bar{C}[k]$ and $M_2[k] = \bar{C}[k])^{-1}(\bar{C}[k]^\prime\bar{C}[k])$ and $M_3[k] = (I-\bar{C}[k] (\bar{C}[k]^\prime\bar{C}[k])^{-1}\bar{C}[k]^\prime)$: 
\begin{eqnarray*}
& & \Scale[0.9]{\left(\bar{c}_i^{[k]} +\epsilon_i\right)^{\prime} \hat{\Sigma}_{k}^{-1}\left(\bar{c}_i^{[k]} +\epsilon_i\right) }\\
&=& \Scale[0.9]{\sigma^{-2} B_k^{\prime} M_3[k] B_k - B_k^{\prime}
\bar{C}[k]M_2[k]^{-1}M_1[k]^{-1}\bar{C}[k]^\prime
B_k}
\end{eqnarray*}
Thus, 
\begin{eqnarray*}
& &    \log \frac{ {Pr}(c_i \mid \hat{\mu}_k, \hat{\Sigma}_k)}{ {Pr}(c_i \mid \hat{\mu}_j, \hat{\Sigma}_j)} \\
& = & -\frac{1}{2} \log \frac{\left|\hat{\Sigma}_{k}\right|}{\left|\hat{\Sigma}_{j}\right|} -\frac{1}{2} \log \frac{\exp (c_i-\hat{\mu}_k)^{\prime} \hat{\Sigma}_{k}^{-1} (c_i-\hat{\mu}_k)}{\exp (c_i-\hat{\mu}_j)^{\prime} \hat{\Sigma}_{j}^{-1} (c_i-\hat{\mu}_j)} \\
& = & 
-\frac{1}{2} \log | \bar{C}[k]\bar{C}[k]^\prime + \sigma^2I |
+ \frac{1}{2} \log | \bar{C}[j]\bar{C}[j]^\prime + \sigma^2I | \\
&  
-&\frac{1}{2} ({\sigma^{-2} B_k^{\prime} M_3[k] B_k + B_k^{\prime}
\bar{C}[k]M_2[k]^{-1}M_1[k]^{-1}\bar{C}[k]^\prime
B_k})\\
& 
+&\frac{1}{2}( {\sigma^{-2} B_j^{\prime} M_3[j] B_j + B_j^{\prime}
\bar{C}[j]M_2[j]^{-1}M_1[j]^{-1}\bar{C}[j]^\prime
B_j} ) \\
& = & 
-\frac{1}{2} \log  | \bar{C}[k]\bar{C}[k]^\prime + \sigma^2I |
+ \frac{1}{2} \log | \bar{C}[j]\bar{C}[j]^\prime + \sigma^2I | \\
& - &
\frac{1}{2} \sigma^{-2} B_k^{\prime} M_3[k] B_k 
+ \frac{1}{2}\sigma^{-2} B_j^{\prime} M_3[j] B_j
\\
& - &
\frac{1}{2} B_k^{\prime}
\bar{C}[k]M_2[k]^{-1}M_1[k]^{-1}\bar{C}[k]^\prime
B_k   \\
& + &
\frac{1}{2}B_j^{\prime}
\bar{C}[j]M_2[j]^{-1}M_1[j]^{-1}\bar{C}[j]^\prime
B_j  
\end{eqnarray*}

\begin{eqnarray*}
& &    \log \frac{ {Pr}(c_i \mid \hat{\mu}_k, \hat{\Sigma}_k)}{ {Pr}(c_i \mid \hat{\mu}_j, \hat{\Sigma}_j)} \\
& \geq & -\frac{1}{2} \sup \log  | \bar{C}[k]\bar{C}[k]^\prime + \sigma^2I | \\
& - & \frac{1}{2}\sigma^{-2}  \sup |B_k^{\prime} M_3[k] B_k - B_k^{\prime} M_3^*[k] B_k | \\
& - & \frac{1}{2}\sigma^{-2}  \sup | B_j^{\prime} M_3[j] B_j - B_j^{\prime} M_3^*[j] B_j |  \\
& - &
\frac{1}{2} \sup B_k^{\prime}
\bar{C}[k]M_2[k]^{-1}M_1[k]^{-1}\bar{C}[k]^\prime
B_k \\
& + &
\frac{1}{2}B_j^{\prime}
\bar{C}[j]M_2[j]^{-1}M_1[j]^{-1}\bar{C}[j]^\prime
B_j \\
& \geq & \frac{1}{2}B_j^{\prime}
\bar{C}[j]M_2[j]^{-1}M_1[j]^{-1}\bar{C}[j]^\prime
B_j - M
\end{eqnarray*}

\noindent Thus, 
\begin{eqnarray*}
{Pr}(\frac{1}{2}B_j^{\prime}
\bar{C}[j]M_2[j]^{-1}M_1[j]^{-1}\bar{C}[j]^\prime
B_j \leq M + m_R) \rightarrow 0
\end{eqnarray*}

If $n \rightarrow \infty$, then the number of samples for each group $n_k = n \times Pr(Z=k) \rightarrow \infty$ and we have $\hat{\mu}_{k} \xrightarrow{p} {\mu}_{k}$ and $\hat{\Sigma}_{k} \xrightarrow{p} {\Sigma}_{k}$ for each $k=1,2,\cdots,K$. If the dimension of representations is sufficiently large and the distribution of representations for different groups can be separated with high probability\footnote{This implicitly assumes that $m_X \rightarrow \infty$. Only when the information in the raw data is sufficient, we can obtain as many representations as possible that nonzero contribute ($\alpha_k$) for all $k$, i.e., $m_R \rightarrow \infty$.}, then  $\left|\hat{\gamma}_{i k}-1_{z_{i}=k}\right|=o_{p}\left(\exp(-({m_R}+M)) \right)$ for a sufficiently large $M > 0$. That means that $\hat{\gamma}_{i k} \xrightarrow{p} Pr(z_{i}=k) \xrightarrow{p} 1_{z_{i}=k}$. 

\paragraph{Proof} \textbf{(2)} As $(m_R, n) \rightarrow \infty$, for each $k$ and for a sufficiently large $M > 0$: $\hat{\gamma}_{i k} \xrightarrow{p} Pr(z_{i}=k) \xrightarrow{p} 1_{z_{i}=k}$. 
We sample the sub-group indicator $z_i$ as group IV: 
$\hat{z}_i \sim \text{Disc} ({\hat \gamma}_{i 1},{\hat \gamma}_{i 2},\cdots,{\hat \gamma}_{i K})$. Due to the randomness, it is possible that the estimated source label $\hat{z}_i$ and the true label ${z}_i$ do not match for each sample $\{x_i, t_i, y_i\}$. 

Nevertheless, in probability, the treatment assignment mechanism of the estimated group label is consistent with that of the real label, i.e., $Pr(\hat{Z}, X, T, Y) \xrightarrow{p} Pr(Z, X, T, Y)$. We can view this phenomenon as the fact that when two identical samples swap labels between groups, the treatment assignment mechanism and the joint distribution will remain the same. Therefore, $\hat{z}_i$ still is an asymptotic IV for treatment effect regression. 


\begin{table*}
  \caption{The Mean Squared Error $mean(std)$ of Different Synthetic Settings ($Data$-$K$-$m_X$)}
  \label{tab:setting}
  \centering
  \scalebox{0.70}{
  \begin{tabular}{cc cccc | cccc}
    \toprule
    & & \multicolumn{4}{c}{Linear-2-5} &  \multicolumn{4}{c}{Linear-3-5} \\
    \cmidrule(r){3-6} \cmidrule(r){7-10}
    & & Poly2SLS &  KernelIV &  DeepIV &  AGMM &  Poly2SLS &  KernelIV &  DeepIV &  AGMM \\
    \midrule
    & None & 0.203(0.017)  & 0.292(0.026)  & 0.371(0.019)  & 0.116(0.015) & 0.312(0.034)  & 0.415(0.047)  & 0.492(0.030)  & 0.129(0.018) \\
    \midrule
    \multirow{4}{*}{\makecell[c]{Summary \\ IV}} & UAS & 0.203(0.017)  & 0.292(0.027)  & 0.376(0.020)  & 0.118(0.012) & 0.312(0.034)  & 0.415(0.046)  & 0.485(0.029)  & 0.130(0.017) \\
    & WAS & 0.204(0.017)  & 0.288(0.040)  & 0.368(0.017)  & 0.127(0.020) & 0.314(0.033)  & 0.417(0.057)  & 0.489(0.028)  & 0.186(0.035) \\
    & ModeIV & 0.203(0.017)  & 0.288(0.025)  & 0.368(0.017)  & 0.113(0.016) & 0.312(0.034)  & 0.418(0.042)  & 0.489(0.030)  & 0.130(0.018) \\
    & AutoIV & 12.739(28.272)  & 0.288(0.028)  & 0.372(0.021)  & 0.118(0.017) & $>100$ & 0.416(0.047)  & 0.486(0.033)  & 0.130(0.017) \\
    \midrule
    \multirow{3}{*}{\makecell[c]{Our \\ Method}} & {}GIV$_{KM}$ & 0.059(0.004)  & 0.250(0.023)  & 0.286(0.018)  & 0.086(0.012) & 0.088(0.020)  & 0.317(0.037)  & 0.361(0.026)  & 0.110(0.013) \\
    & {}GIV$_{KM}$* & 0.059(0.004)  & 0.250(0.023)  & 0.289(0.023)  & 0.086(0.012) & 0.142(0.087)  & 0.274(0.031)  & 0.284(0.029)  & 0.089(0.007) \\
    & {}GIV$_{EM}$ & \bf 0.058(0.030)  & \bf 0.141(0.016)  & \bf 0.076(0.006)  & \bf 0.057(0.005) & \bf 0.045(0.004)  & \bf 0.167(0.028)  & \bf 0.104(0.008)  & \bf 0.067(0.007) \\
    \hline
    & TrueIV & \bf 0.044(0.006)  & \bf 0.139(0.016)  & \bf 0.078(0.008)  & \bf 0.058(0.006) & \bf 0.044(0.004)  & \bf 0.169(0.032)  & \bf 0.101(0.010)  & \bf 0.069(0.005) \\
  \bottomrule
  
    & & \multicolumn{4}{c}{Linear-5-5} &  \multicolumn{4}{c}{Linear-3-10} \\
    \cmidrule(r){3-6} \cmidrule(r){7-10}
    & & Poly2SLS &  KernelIV &  DeepIV &  AGMM &  Poly2SLS &  KernelIV &  DeepIV &  AGMM \\
    \midrule
    & None & 0.474(0.044)  & 0.538(0.093)  & 0.658(0.042)  & 0.103(0.022) & 0.292(0.023)  & 0.400(0.096)  & 0.655(0.051)  & 0.076(0.015) \\
    \midrule
    \multirow{4}{*}{\makecell[c]{Summary \\ IV}} & UAS & 0.474(0.044)  & 0.540(0.094)  & 0.651(0.049)  & 0.102(0.020) & 0.292(0.023)  & 0.400(0.096)  & 0.656(0.055)  & 0.077(0.015) \\
    & WAS & 0.478(0.045)  & 0.553(0.100)  & 0.651(0.045)  & 0.108(0.020) & 0.294(0.024)  & 0.413(0.098)  & 0.657(0.048)  & 0.079(0.015) \\
    & ModeIV & 0.474(0.044)  & 0.547(0.094)  & 0.657(0.042)  & 0.103(0.019) & 0.293(0.024)  & 0.403(0.097)  & 0.653(0.062)  & 0.079(0.016) \\
    & AutoIV & $>100$ & 0.540(0.093)  & 0.656(0.055)  & 0.108(0.026) & $>100$ & 0.399(0.096)  & 0.654(0.058)  & 0.077(0.014) \\
    \midrule
    \multirow{3}{*}{\makecell[c]{Our \\ Method}} & {}GIV$_{KM}$ & 0.268(0.012)  & 0.374(0.080)  & 0.424(0.046)  & 0.066(0.010) & 0.098(0.042)  & 0.357(0.076)  & 0.466(0.055)  & 0.073(0.015) \\
    & {}GIV$_{KM}$* & 0.196(0.290)  & 0.231(0.063)  & \bf 0.208(0.022)  & 0.052(0.005) & 0.090(0.028)  & \bf 0.302(0.063)  & 0.258(0.032)  & 0.062(0.010) \\
    & {}GIV$_{EM}$ & \bf 0.192(0.382)  & \bf 0.207(0.042)  & \bf 0.146(0.032)  & \bf 0.046(0.007) & \bf 0.062(0.096)  & \bf 0.302(0.061)  & \bf 0.258(0.033)  & \bf 0.051(0.011) \\
    \hline
    & TrueIV & \bf 0.130(0.144)  & \bf 0.199(0.041)  & 0.239(0.040)  & \bf 0.043(0.006) & \bf 0.032(0.003)  & 0.311(0.059)  & \bf 0.102(0.016)  & \bf 0.054(0.012) \\
  \bottomrule
  \end{tabular}
  }
\end{table*}


\subsection{Asymptotic Results of ITE Estimation}
\paragraph{Identification of ITE} 
Under \emph{the additive noise assumption \ref{ass:noise}}, the identification results for ITE  $g(\cdot)$ were developed by \citep{Identify3:newey2003instrumental,IVMethod1:hartford2017deepiv}. Therefore, we plug {}GIV into IV regression methods to estimate ITE  $g(\cdot)$. 
\begin{theorem} \textbf{Asymptotic Results of ITE Estimation.}
Taking the expectation of outcome function in Eq. (\ref{eq:Y}) conditional on $\{Z,X\}$ and applying the above assumptions, we establish the relationship:
\begin{eqnarray}
\label{eq:inverse}
\nonumber
    \mathbb{E}[Y \mid Z,X] = \mathbb{E}[g(T,X) \mid Z,X] + \mathbb{E}[\epsilon_Y\mid X] \\ = \int [g(T,X) + C] dF(T \mid Z,X),
\end{eqnarray}
where $dF(T \mid Z,X)$ is the conditional treatment distribution, $C$ is constant as $T$ is changed. 
The relationship in Eq. (\ref{eq:inverse}) defines an inverse problem for $g(\cdot)$ in terms of two directly observable functions: $\mathbb{E}[Y \mid Z,X]$ and $dF(T \mid Z,X)$.
\end{theorem}

\paragraph{Proof}

In the non-linear scenario, the relationship between the outcome process and reduced form belongs a 1st Fredholm integral equation and leads an ill-posed inverse problem \citep{Identify3:newey2003instrumental}. 
Considering the identification of a general outcome model in Eq. (\ref{eq:Y}):
\begin{eqnarray}
\nonumber
Y = g(X, T) + \epsilon_Y, \mathbb{E}[\epsilon_Y \mid Z] = \mathbb{E}[\epsilon_Y] = 0.
\label{eq:noise}
\end{eqnarray}
where $g(\cdot)$ denotes a true, unknown structural function of interest. For a consistency estimation, \citep{Identify3:newey2003instrumental,IVMethod1:hartford2017deepiv} identified the causal effect as the solution of an integral equation: 
\begin{eqnarray}
\mathbb{E}[Y \mid Z, X] & = & \mathbb{E}[g(X, T) \mid Z, X] + \mathbb{E}[\epsilon_Y \mid X] \nonumber \\
& = & \int \left[g(T,X) + \mathbb{E}[\epsilon_Y \mid X]\right] dF(T \mid Z,X) \nonumber \\
& = & \int \left[g(T,X) + C\right] dF(T \mid Z,X) \nonumber \\
& = & \int \hat{g}(T,X) dF(T \mid Z,X)
\end{eqnarray}
where $F$ denotes the conditional cumulative distribution function of $T$ given $\{Z,X\}$, and $\hat{g}(T,X)=g(T,X) + C$. Given two observable functions $\mathbb{E}[Y \mid Z, X]$ and $F(T \mid Z,X)$, $\hat{g}(T,X)$ is the solution of the inverse problem. 
Therefore, we characterize the identification of structural functions as completeness of certain conditional distributions $\mathbb{E}[\epsilon_Y \mid Z] = 0$. 

In the parametric/nonparametric model (Eq. (\ref{eq:noise})), the identification/uniqueness of $\hat{g}(T,X)$ is equivalent to the nonexistence of any function $\delta(X,T): = \hat{g}(T,X) - g(T,X) \not = 0$ such that $\mathbb{E}[\delta(X,T) \mid Z] = 0$. Plugging the LatGIV into IV-based methods, we can predict ITE under assumption \ref{ass:noise} and $C=\mathbb{E}[Y-\hat{g}(T,X)]$.

\begin{figure*}[t]
    \centering
    \includegraphics[width=1.0\textwidth]{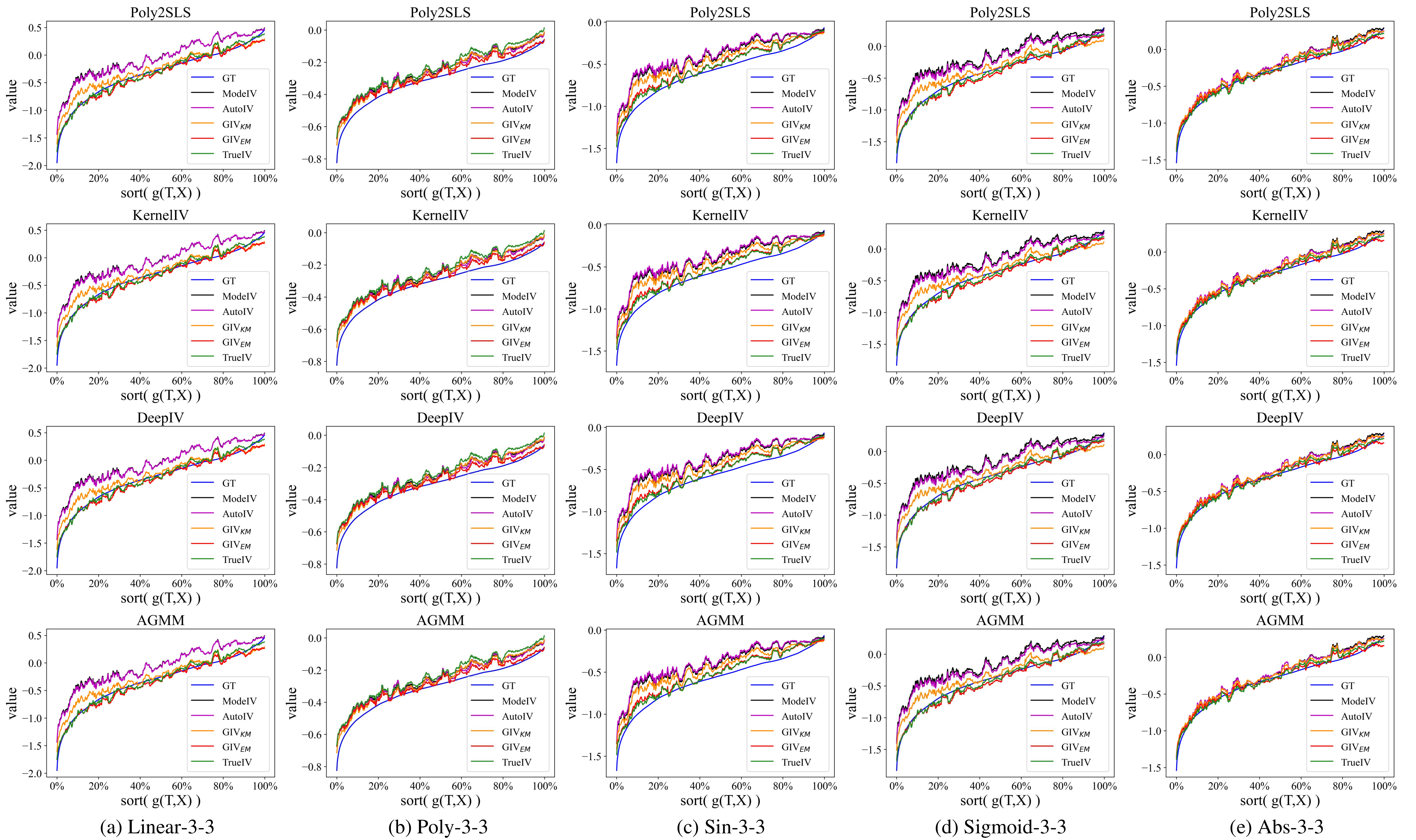}
    \caption{Treatment Effect Estimation (sorted by Ground-Truth $g(T,X)$) in Different Scenario $Data$-$3$-$3$.}
    \label{fig:GIVplot-5}
\end{figure*}

\section{The Experiments for Stability}
\label{sec:stablity}

We increase the critical level of simulation and set \textbf{$Data$-$K$-$m_X$} with different group numbers $K$ and dimensions of covariates $m_X$ to test the stability of our {}GIV. 
Comparing with the results of setting Linear-3-3 in Table \ref{tab:linear}, Linear-3-5 and Linear-3-10 in Table \ref{tab:setting}, {}GIV$_{EM}$ consistently achieves Top2 performance as the dimensions of covariates change. Adjusting the number of latent groups (Linear-2-5,Linear-3-5,Linear-5-5 in Table \ref{tab:setting}), {}GIV$_{EM}$ also shows in stability and is in Top2. 
In the above settings, {}GIV$_{EM}$ has outstanding performance, which is close to TrueIV. 

\begin{table*}
  \caption{The Full Results of MSE $mean(std)$ of IHDP \& PM-CMR Dataset}
  \label{tab:True}
  \centering
  \scalebox{0.65}{
  \begin{tabular}{ccccccccccc}
    \toprule
    \multicolumn{11}{c}{IHDP Dataset} \\
    \hline 
    & & Poly2SLS &  NN2SLS &  KernelIV &  DualIV &  DeepIV &  OneSIV &  DFIV &  DeepGMM &  AGMM \\
    \midrule
    & NoneIV & 0.238(0.132)  & 2.127(1.967)  & 0.456(0.243)  & 0.762(0.279)  & 0.583(0.240)  & 0.444(0.186)  & 0.979(0.391)  & 0.239(0.186)  & 0.140(0.063) \\
    \midrule
    \multirow{4}{*}{\makecell[c]{Summary \\ IV}} & UAS & 0.238(0.133)  & 2.712(2.795)  & 0.457(0.244)  & 0.659(0.254)  & 0.574(0.245)  & 0.447(0.185)  & 1.024(0.426)  & 0.243(0.129)  & 0.142(0.061) \\
    & WAS & 0.239(0.134)  & 1.954(1.836)  & 0.455(0.241)  & 0.746(0.254)  & 0.567(0.226)  & 0.456(0.185)  & 1.010(0.408)  & 0.195(0.102)  & 0.144(0.059) \\
    & ModeIV & 0.240(0.133)  & 2.511(2.048)  & 0.460(0.246)  & 0.752(0.277)  & 0.572(0.245)  & 0.468(0.193)  & 0.989(0.387)  & 0.220(0.085)  & 0.149(0.060) \\
    & AutoIV & $>100$ & 2.392(2.052)  & 0.457(0.243)  & 0.682(0.257)  & 0.583(0.250)  & 0.458(0.199)  & 0.998(0.405)  & 0.196(0.087)  & 0.142(0.069) \\
    \midrule
    \multirow{2}{*}{\makecell[c]{Our Method}} & {}GIV$_{KM}$ & 0.078(0.029)  & 1.009(0.964)  & 0.354(0.179)  & \bf 0.651(0.266)  & 0.505(0.203)  & 0.383(0.155)  & 0.967(0.343)  & \bf 0.131(0.037)  & 0.112(0.050) \\
    & {}GIV$_{EM}$ & \bf 0.034(0.011)  & \bf 0.585(1.342)  & \bf 0.202(0.173)  & \bf 0.653(0.255)  & \bf 0.482(0.228)  & \bf 0.283(0.163)  & \bf 0.967(0.333)  & \bf 0.137(0.041)  & \bf 0.095(0.035) \\
    \hline
    & TrueIV & \bf 0.033(0.009)  & \bf 0.146(0.044)  & \bf 0.151(0.060)  & 0.654(0.256)  & \bf 0.458(0.166)  & \bf 0.227(0.079)  & \bf 0.948(0.342)  & 0.152(0.040)  & \bf 0.093(0.035) \\
  \bottomrule
  \toprule
    \multicolumn{11}{c}{PM-CMR Dataset} \\
    \hline 
    & & Poly2SLS &  NN2SLS &  KernelIV &  DualIV &  DeepIV &  OneSIV &  DFIV &  DeepGMM &  AGMM \\
    \midrule
    & NoneIV & 0.181(0.044)  & 1.241(0.646)  & 0.352(0.198)  & 0.727(0.230)  & 0.409(0.160)  & 0.369(0.182)  & 0.995(0.166)  & 0.145(0.043)  & 0.130(0.064) \\
    \midrule
    \multirow{4}{*}{\makecell[c]{Summary \\ IV}} & UAS & 0.181(0.044)  & 1.439(0.733)  & 0.352(0.198)  & 0.656(0.205)  & 0.404(0.164)  & 0.365(0.181)  & 0.994(0.177)  & 0.213(0.096)  & 0.128(0.064) \\
    & WAS & 0.181(0.044)  & 0.939(0.465)  & 0.372(0.207)  & 0.906(0.251)  & 0.417(0.164)  & 0.417(0.166)  & \bf 0.967(0.202)  & 0.200(0.088)  & 0.157(0.080) \\
    & ModeIV & 0.181(0.044)  & 1.515(0.687)  & 0.359(0.201)  & 0.749(0.203)  & 0.406(0.150)  & 0.391(0.190)  & 1.035(0.186)  & 0.204(0.108)  & 0.131(0.070) \\
    & AutoIV & 0.179(0.044)  & 1.224(0.719)  & 0.351(0.198)  & 0.706(0.220)  & 0.409(0.180)  & 0.379(0.204)  & 0.984(0.195)  & 0.227(0.193)  & 0.129(0.064) \\
    \midrule
    \multirow{2}{*}{\makecell[c]{Our Method}} & {}GIV$_{KM}$ & 0.088(0.044)  & 0.719(0.452)  & 0.329(0.202)  & \bf 0.658(0.208)  & 0.381(0.165)  & 0.341(0.178)  & 1.049(0.199)  & 0.174(0.062)  & 0.117(0.053) \\
    & {}GIV$_{EM}$ & \bf 0.048(0.012)  & \bf 0.624(0.395)  & \bf 0.308(0.210)  & \bf 0.666(0.203)  & \bf 0.339(0.184)  & \bf 0.306(0.164)  & \bf 0.982(0.205)  & \bf 0.125(0.052)  & \bf 0.085(0.045) \\
    \hline
    & TrueIV & \bf 0.028(0.007)  & \bf 0.190(0.054)  & \bf 0.140(0.074)  & 0.678(0.202)  & \bf 0.141(0.054)  & \bf 0.154(0.071)  & 0.993(0.207)  & \bf 0.112(0.048)  & \bf 0.054(0.023) \\
  \bottomrule
  \end{tabular}
  }
\end{table*}

\section{The Experiments for Non-linear Cases}
\label{sec:NonExp}

To verify the effectiveness of {}GIV in non-linear cases, we design 5 different treatment functions $f_X(\cdot)$ to evaluate the treatment effect estimation performance of Meta-EM algorithm. We select the SOTA IV-based methods (Poly2SLS, KernelIV, DeepIV, AGMM) in four lines to evaluate {}GIV. We plot the estimated value of effect function with T=do(t) and sort it by Ground-Truth (GT) for different synthetic scenarios. The results (Fig. \ref{fig:GIVplot-5}) show {}GIVs (with Meta-KM or Meta-EM) achieve SOTA performance, especially {}GIV$_{EM}$ achieves comparable results with TrueIV and estimated 
outcome curves from {}GIV$_{EM}$ approximate the true curve.

\section{The Full Results in IHDP \& PM-CMR}
\label{sec:fullExp}


\begin{figure}
    \centering
    \includegraphics[scale=0.35]{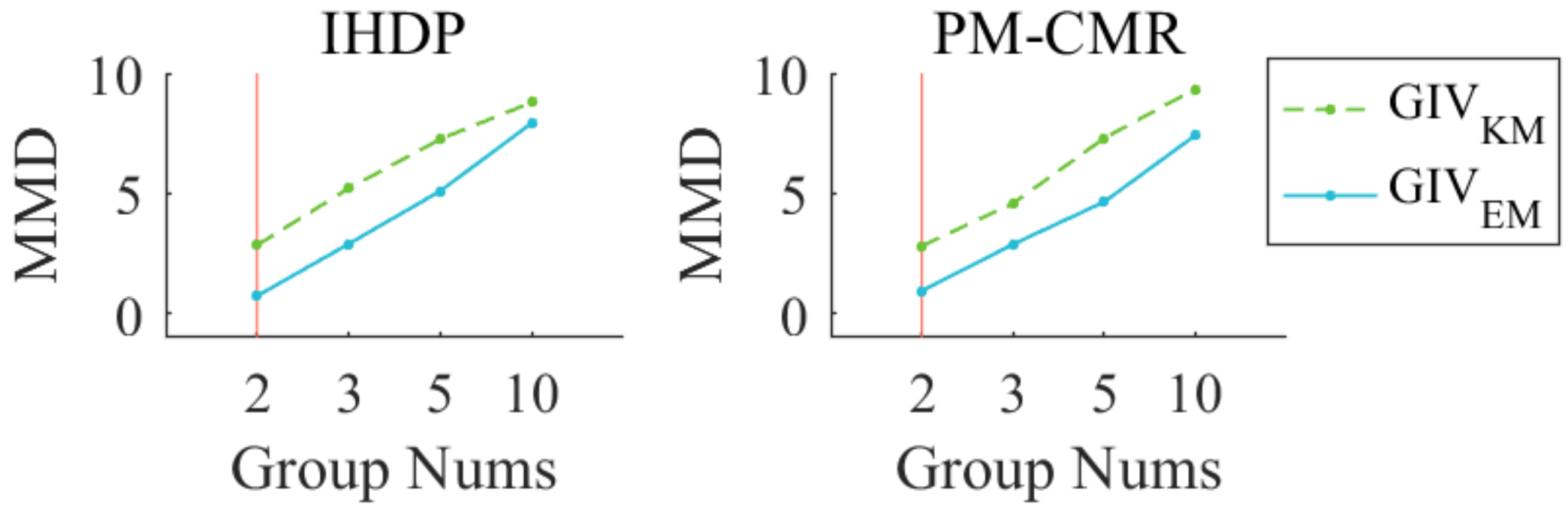}
    \caption{MMD for Selection of Group Number in IHDP \& PM-CMR Dataset.}
    \label{fig:GIVRealMMD}
\end{figure}



\begin{figure}
    \centering
    \includegraphics[scale=0.35]{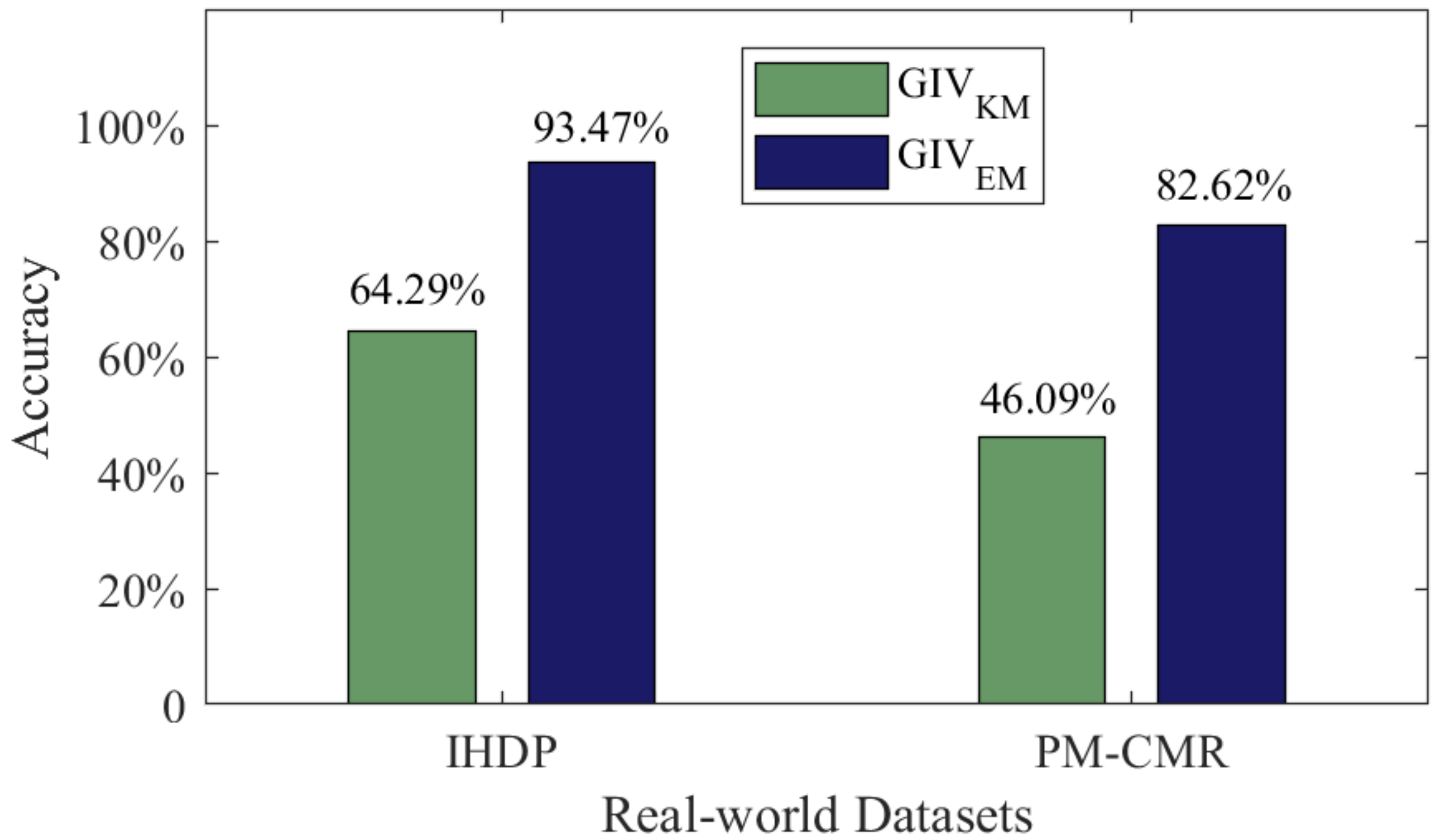}
    \caption{Reconstruction Accuracy of the Group IV in IHDP \& PM-CMR Dataset.}
    \label{fig:GIVRealACC}
\end{figure}


\subsection{Real-world datasets}
\label{sec:realData}

Similar to previous methods\citep{ConfMethod10:nie2020vcnet,IVMethod1:hartford2017deepiv,ConfMethod11:bica2020scigan,ConfMethod9:schwab2020drnet}, we perform experiments on two semi-synthetic real-world datasets \textbf{IHDP} \citep{ConfMethod2:shalit2017cfr} \& \textbf{PM-CMR} \citep{ConfMethod13:wyatt2020dataset3}, as the counterfactual outcomes are rarely available for real-world data. 
Both two datasets are randomly split into training (63\%), validation (27\%) and testing (10\%). We perform 10 replications to report the mean squared error (MSE) and its standard deviations (std) of the treatment effect estimation.

\textbf{IHDP}
\citep{ConfMethod2:shalit2017cfr}. The Infant Health and Development Program ({IHDP}) comprises 747 units with 6 pre-treatment continuous variables and 19 discrete variables related to the children and their mothers, aiming at evaluating the effect of specialist home visits on the future cognitive test scores of premature infants. 
From the original data, We select all 6 continuous variables as the confounders to replace the covariates $X$ in Eq. (\ref{eq:treatment})\&(\ref{eq:outcome}) to generate the treatment $T$ and corresponding outcome $Y$. 

\textbf{PM-CMR}
\citep{ConfMethod13:wyatt2020dataset3}. The PM-CMR study the impact of $PM_{2.5}$ partical level on the cardiovascular mortality rate (CMR) in 2132 counties in the US using the data provided by the National Studies on Air Pollution and Health \citep{ConfMethod13:wyatt2020dataset3}.
Then we use 6 continuous variables about CMR in each city as the confounders to replace the covariates $X$ in Eq. (\ref{eq:treatment})\&(\ref{eq:outcome}) to generate the treatment $T$ and the corresponding outcome $Y$. 

\subsection{Results}
By estimating the latent differentiated covariate-treatment distribution parameters across groups, Meta-EM reconstructs the latent IV. From the results in Fig. \ref{fig:GIVRealACC}) \& \ref{fig:GIVRealMMD}, we have the following observation: (1) the optimal parameter ($K=2$) identified by Meta-EM is consistent with the ground truth, meaning Meta-EM can find the optimal parameter $K$; (2) the MMD of Meta-EM is still significantly smaller than that of Meta-KM and (3) the reconstruction accuracy reaches 93.47\% and 82.62\% on \textbf{IHDP} and \textbf{PM-CMR}, however, K-Means is only 64.29\% and 46.09\%. (2-3) demonstrate Meta-EM can automatically find the optimal IV, but K-Means cannot.

To verify that {}GIV$_{EM}$ with higher reconstruction accuracy achieves better performance to predict treatment effect, we assess {}GIV and Summary IVs' performance in treatment effect estimation with the covariates from the real-world data IHDP \& PM-CMR. We perform 10 replications and report the mean and standard deviations of MSE in the treatment effect estimation here. The full results of MSE $mean(std)$ of IHDP \& PM-CMR Dataset with T=do(t) are shown in Table \ref{tab:True}, 
{}GIV$_{EM}$ shows consistent and robust performance, always maintaining the performance of top-2 and almost achieving the same effect as TrueIV on IHDP \& PM-CMR Datasets. Compared with {}GIV$_{EM}$, the performance of {}GIV$_{KM}$ exceeds most baselines in downstream tasks, but it is still inferior to {}GIV$_{EM}$ and TrueIV.

\end{document}